\documentclass[10pt,twocolumn,letterpaper]{article}
\usepackage[pagenumbers]{cvpr} 
\usepackage{amsmath}
\usepackage{amssymb}
\usepackage{multirow}
\usepackage{graphicx}
\usepackage{subcaption}
\usepackage{bbm}
\usepackage{bm}
\usepackage{array}
\usepackage{makecell}
\usepackage{cuted}
\usepackage{capt-of}
\PassOptionsToPackage{table}{xcolor}
\usepackage{colortbl}

\definecolor{cvprblue}{rgb}{0.21,0.49,0.74}
\usepackage[pagebackref,breaklinks,colorlinks,allcolors=cvprblue]{hyperref}

\title{U-Know-DiffPAN: An Uncertainty-aware Knowledge Distillation Diffusion Framework with Details Enhancement for PAN-Sharpening}

\author{Sungpyo Kim\\
[0.3em]
KAIST\\
{\tt\footnotesize ksp04204@kaist.ac.kr}
\and
Jeonghyeok Do\\
[0.3em]
KAIST\\
{\tt\footnotesize ehwjdgur0913@kaist.ac.kr}
\and
Jaehyup Lee \footnotemark[2]\\
[0.3em]
KNU\\
{\tt\footnotesize jaehyuplee@knu.ac.kr}
\and
Munchurl Kim \footnotemark[2]\\
[0.3em]
KAIST\\
{\tt\footnotesize mkimee@kaist.ac.kr}
\and
\small{\url{https://kaist-viclab.github.io/U-Know-DiffPAN-site}}
}

\begin{document}
\maketitle

\begin{strip}\centering
    \vspace{-1.7cm}
    \includegraphics[width=1\linewidth]{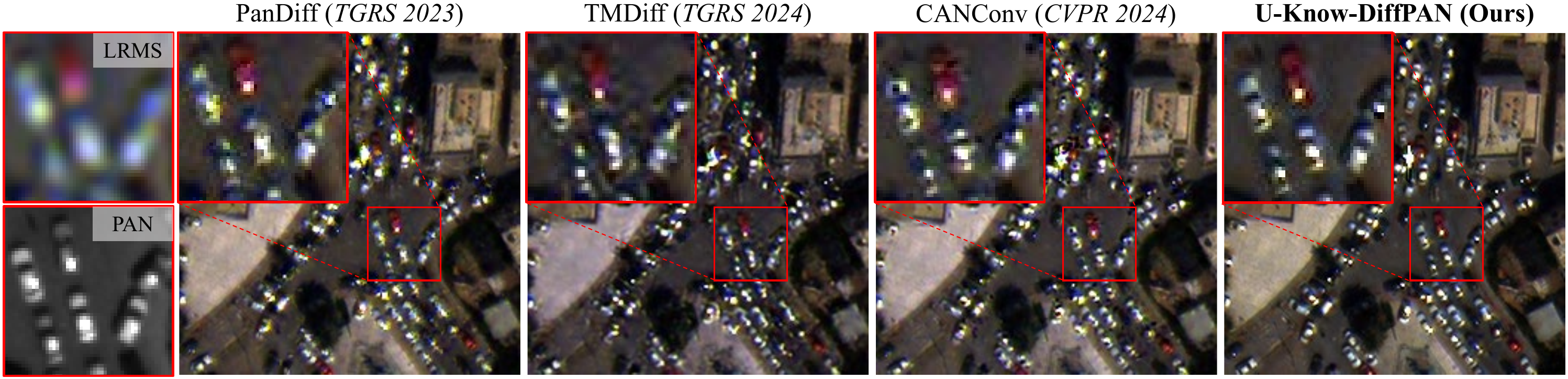}
    \vspace{-0.6cm}
    \captionof{figure}{Visual comparison of PAN-sharpening results on the full-resolution WV3 dataset. The rightmost image shows the output of our proposed U-Know-DiffPAN framework, specifically the result produced by FSA-S (frequency selective attention student network). Notably, the proposed framework generates more detailed and robust results, particularly in high-uncertainty regions, outperforming the state-of-the-art model CANConv \cite{duan2024content} and recent diffusion-based methods PanDiff \cite{meng2023pandiff} and TMDiff \cite{xing2024empower}. As highlighted in the red box, our approach successfully restores challenging highly-uncertain regions, such as cars, where other models fall short.}
    \label{fig:first}
\end{strip}

{
  \renewcommand{\thefootnote}%
    {\fnsymbol{footnote}}
  \footnotetext[2]{Co-corresponding authors (equal advising).}
}

\begin{abstract}
Conventional methods for PAN-sharpening often struggle to restore fine details due to limitations in leveraging high-frequency information. Moreover, diffusion-based approaches lack sufficient conditioning to fully utilize Panchromatic (PAN) images and low-resolution multispectral (LRMS) inputs effectively. To address these challenges, we propose an uncertainty-aware knowledge distillation diffusion framework with details enhancement for PAN-sharpening, called U-Know-DiffPAN. The U-Know-DiffPAN incorporates uncertainty-aware knowledge distillation for effective transfer of feature details from our teacher model to a student one. The teacher model in our U-Know-DiffPAN captures frequency details through freqeuncy selective attention, facilitating accurate reverse process learning. By conditioning the encoder on compact vector representations of PAN and LRMS and the decoder on Wavelet transforms, we enable rich frequency utilization. So, the high-capacity teacher model distills frequency-rich features into a lightweight student model aided by an uncertainty map. From this, the teacher model can guide the student model to focus on difficult image regions for PAN-sharpening via the usage of the uncertainty map. Extensive experiments on diverse datasets demonstrate the robustness and superior performance of our U-Know-DiffPAN over very recent state-of-the-art PAN-sharpening methods.
\end{abstract}   
\section{Introduction}
\label{sec:intro}
Satellite imagery plays a vital role in a wide range of applications, including environmental monitoring, surveillance, and mapping \cite{zhang2024anisotropic, amro2011survey, loncan2015hyperspectral, duran2017survey, ciotola2024hyperspectral}. High-resolution multispectral (HRMS) images that combine high spatial and spectral fidelity are commonly generated using PAN-sharpening techniques, fusing low-resolution multispectral (LRMS) images with high-resolution (HR) panchromatic (PAN) images. The primary objective of PAN-sharpening is to retain the spatial detail of PAN images while preserving the spectral information of LRMS images. Recently, deep learning (DL)-based approaches have gained attraction in PAN-sharpening tasks, leveraging deep networks to perform end-to-end image regression. However, DL models are typically limited by the local receptive field of convolutional layers, which primarily focus on local patterns and fail to capture long-range dependencies. This limitation often leads to overly smooth outputs with reduced high-frequency details.

To overcome this limitation, generative models such as generative adversarial networks (GANs) \cite{goodfellow2014generative} have been applied to PAN-sharpening, aiming to produce more realistic results. Despite their success, GANs \cite{liu2023mun, ma2020pan} pose additional challenges, including training instability, mode collapse, and artifact generation, which can degrade image quality. More recently, diffusion models \cite{ho2020denoising, song2020denoising, xia2023diffir, saharia2022image} have demonstrated strong performance across a variety of image generation and restoration tasks. In PAN-sharpening, diffusion-based methods \cite{meng2023pandiff, xing2024empower} have shown promise by using PAN and LRMS images as conditioning inputs. However, they are limited by simple conditioning strategies that do not fully utilize the complementary information provided by the PAN and LRMS inputs, thus leading to suboptimal restoration quality. Additionally, diffusion models often incur high computational costs and longer inference times compared to other DL-based methods, limiting their practical applicability.

To address these challenges, we propose U-Know-DiffPAN demonstrated in Fig.~\ref{Method/kd_model}, a novel PAN-sharpening framework that incorporates an uncertainty-aware knowledge distillation (U-Know) strategy. Our framework is designed to fully leverage the frequency information in PAN and LRMS inputs while reducing computational costs. Our teacher model, is built to maximize information extraction from PAN and LRMS images by utilizing (i) a Feed Forward Attention (FFA) block in the encoder to capture compact vector representations of PAN and LRMS, and (ii) a High Quality Frequency Enhancement (HQFE) block in the decoder, composed of Fourier Transform Channel Attention (FTCA) and Stationary Wavelet Transform Cross Attention (SWTCA) modules. The FTCA enhances crucial frequency components through Fourier transform-based \cite{weinstein1971data} attention, while the SWTCA incorporates high-frequency information from PAN and low-frequency information from LRMS into the decoder.

Our U-Know strategy leverages uncertainty-aware knowledge distillation by transferring the features from the teacher model to a light-weight student model. The teacher model denoises HRMS images along with an uncertainty map, identifying spatial regions where the prediction confidence is low. Then based of the U-Know strategy, the teacher model distills to the student model not only the frequency-aware features but also the uncertainty map to address spatial weaknesses in the teacher’s predictions. This approach leads to improved performance and reduced computational cost, making our U-Know-DiffPAN more effective and efficient for PAN-sharpening. Our contributions are summarized as follows:

\begin{itemize}
    \item We \textit{firstly} incorporate the uncertainty-aware knowledge distillation into diffusion based PAN-sharpening with the guidance of pixel-wise restoration uncertainty and frequency-selective attention for details enhancement; 
    \item Our teacher model leverages compact vector representations to enhance encoding efficiency and employs frequency selective attention in the decoder to refine frequency components for PAN-sharpening.
    \item We conduct extensive experiments on multiple datasets \cite{deng2022machine}, including WV3, QB, and GF2, demonstrating state-of-the-art performance and surpassing very recent methods in PAN-sharpening with \textit{large margins}.
\end{itemize}    
\section{Related Works}
\subsection{PAN-Sharpening}
PAN-sharpening is a technique to fuse high-resolution (HR) panchromatic (PAN) images with low-resolution multispectral (LRMS) images to produce HR multispectral (HRMS) images. Traditional PAN-sharpening techniques are broadly categorized into three approaches: component substitution (CS) \cite{choi2010new, shensa1992discrete}, multi-resolution analysis (MRA) \cite{otazu2005introduction, aiazzi2006mtf}, and variational optimization (VO) \cite{ballester2006variational}. Modern PAN-sharpening methods primarily consist of CNN-based \cite{yang2017pannet, wald1997fusion, jin2022lagconv, duan2024content}, transformer-based \cite{jin2022lagconv, zhang2024dcpnet}, and GAN-based \cite{ma2020pan, liu2023mun} models. LAGConv \cite{jin2022lagconv} introduces local-context adaptive convolutional kernels with a global harmonic bias mechanism. S2DBPN \cite{zhang2023spatial} uses a spatial-spectral dual back-projection network for effective PAN and LRMS fusion. DCPNet \cite{zhang2024dcpnet} establishes a transformer-based dual-task parallel framework to optimize reconstruction. CANConv \cite{duan2024content} integrates a non-local adaptive convolution module, enhancing spatial adaptability and non-local self-similarity. Besides the methods cited above, there are many other DL-based methods for PAN-sharpening \cite{huang2015new, ciotola2022pansharpening, hu2020pan, gastineau2020residual, doi2019sscnet}. While deep learning (DL)-based methods can capture complex mappings, they encounter challenges. CNN-based approaches tend to produce overly smooth results due to local receptive field of convolution layers. GAN-based models, though capable of generating realistic textures, may suffer from instability and introduce unrealistic artifacts. 

\noindent\textbf{Diffusion-based approaches.}\quad Diffusion models (DMs) \cite{ho2020denoising, song2020denoising} offer a promising alternative to conventional DL-based methods for image restoration \cite{xia2023diffir, wang2024exploiting}, iteratively refining noisy images through progressive denoising. Unlike GANs \cite{goodfellow2014generative, karras2019style}, which can be prone to instability, DMs use iterative denoising steps with flexible conditioning, making them robust for tasks such as image restoration and PAN-sharpening. Recent diffusion-based PAN-sharpening models \cite{meng2023pandiff, xing2024empower} have shown significant advancements. PanDiff \cite{meng2023pandiff}, however, employs a relatively simplistic conditioning approach that limits its ability to fully utilize the information from the input data, sometimes resulting in suboptimal restoration quality. In contrast, TMDiff \cite{xing2024empower} employs a complex conditioning mechanism, requiring an additional pretrained CLIP \cite{radford2021clip} text encoder, which adds unnecessary complexity without substantial performance gains. Our proposed framework, U-Know-DiffPAN, handles these limitations by integrating an advanced conditioning strategy that maximizes the use of input information. Specifically, our U-Know-DiffPAN conditions the encoder on compact vector representations of PAN and LRMS, while the decoder is conditioned on the stationary wavelet transform \cite{nason1995stationary} of PAN and LRMS. This approach allows our U-Know-DiffPAN to effectively capture both frequency details and spatial fidelity, achieving superior restoration quality compared to existing diffusion-based methods.

\subsection{Knowledge Distillation}
Knowledge Distillation (KD) \cite{hinton2015distilling} is a well-known method for transferring knowledge from a large, model (teacher) to a smaller model (student), enabling deployment in resource-constrained settings while retaining performance. KD has demonstrated effectiveness across a range of computer vision tasks, including classification \cite{zhang2021data, xu2020feature, chen2019knowledge}, object detection \cite{wang2019distilling, yao2021g, huang2020comprehensive, zheng2021se}, and image restoration \cite{yang2024vitkd, zhang2023data, sun2021learning}. However, previous KD methods, often fall short in tasks such as super-resolution \cite{zhang2023data}, where capturing complicated detail is essential. This limitation arises because conventional KD student learns all regions equally, without focusing on areas that require concentrated learning. To address this, we propose uncertainty-aware knowledge distillation (U-Know), which introduces an uncertainty map into the distillation process. Our U-Know strategy informs the student of uncertain regions, areas the teacher model finds challenging to restore, enabling the student model to refine crucial details, thereby enhancing fidelity in both spatial and spectral domains.
\section{Methods}
\subsection{Overview of U-Know-DiffPAN}
\begin{figure}[tbp]
    \centering{\includegraphics[width=\columnwidth]{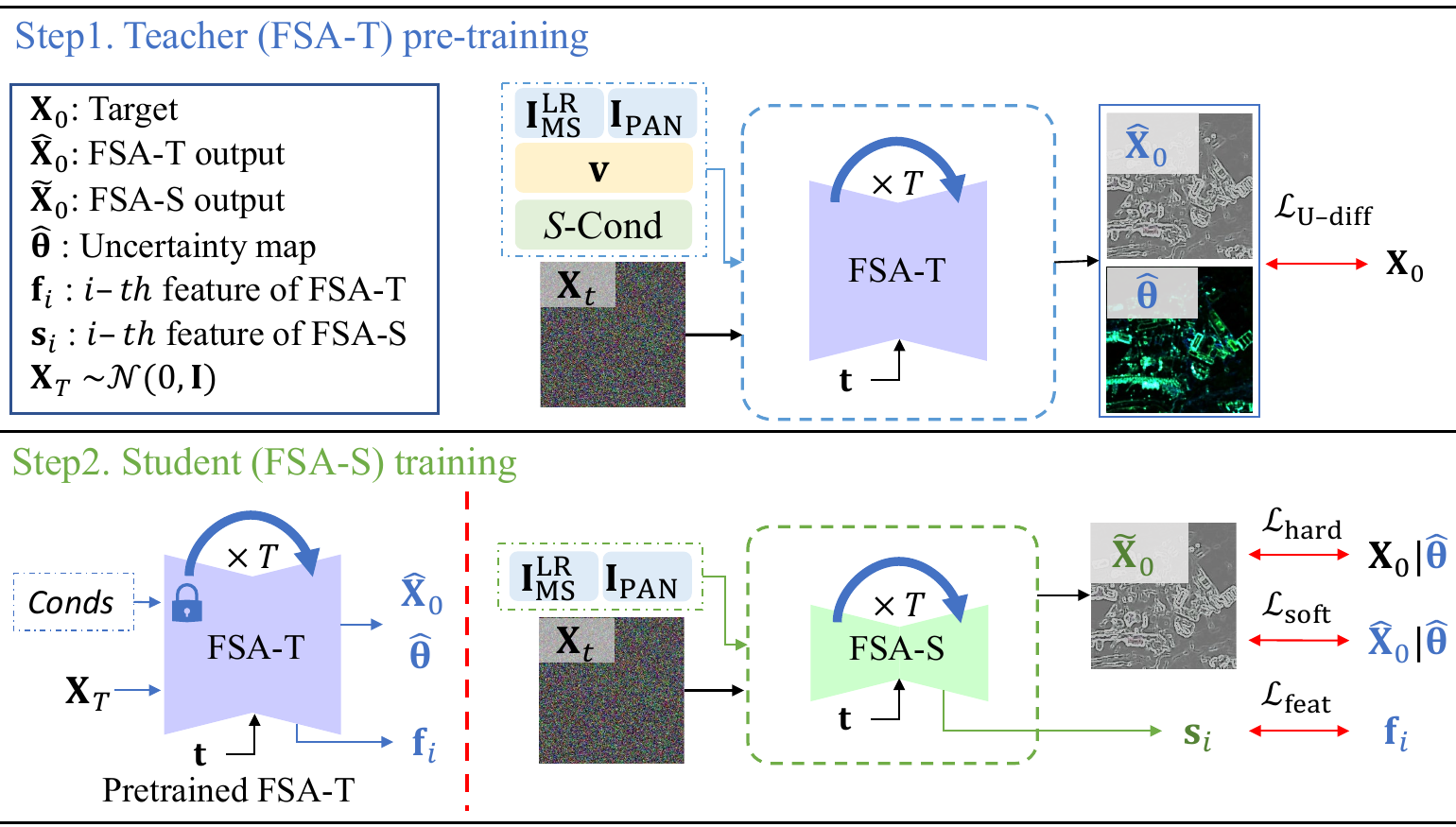}}
    \vspace{-0.5cm}
    \caption{Overview of our uncertainty-aware knowledge-distillation diffusion framework with details enhancement, called U-Know-DiffPAN.}
    \label{Method/kd_model}
    \vspace{-0.4cm}
\end{figure}

Fig.~\ref{Method/kd_model} shows the overview of our U-Know-DiffPAN framework. The proposed U-Know-DiffPAN framework comprises a teacher network with frequency-selective attention (FSA-T) and a student network (FSA-S), designed within a knowledge distillation paradigm. PAN image $\mathbf{I}_\text{PAN}$ and interpolated LRMS image $\mathbf{I}_\text{MS}^\text{LR}$ are inputted in concatenation to generate HRMS image $\mathbf{I}_\text{MS}^\text{HR}$. Both FSA-T and FSA-S are designed to predict the residuals as:
\begin{equation}
    \mathbf{X}_{0} = \mathbf{I}_\text{MS}^\text{HR} - \mathbf{I}_\text{MS}^\text{LR}.
\end{equation}
Our framework operates in two main stages: (i) pre-training the FSA-T within a diffusion process to produce an initial prediction $\widehat{\mathbf{X}}_0$ alongside an uncertainty map $\hat{\bm\theta}$ that identifies spatially weak regions; and (ii) training the FSA-S by leveraging this uncertainty map to guide the FSA-S in refining these regions through the KD. In Sec.~\ref{sec:32}, we describe the FSA-T’s diffusion-based training process, which incorporates frequency information from PAN and LRMS inputs. Sec.~\ref{sec:33} details the FSA-T architecture, specifically designed to enhance frequency components. Finally, in Sec.~\ref{sec:34}, we explain the FSA-S’s uncertainty-aware knowledge distillation framework, which utilizes the uncertainty map to achieve precise and reliable PAN-sharpening.

\begin{figure*}[tbp]
    \centering{\includegraphics[width=\textwidth]{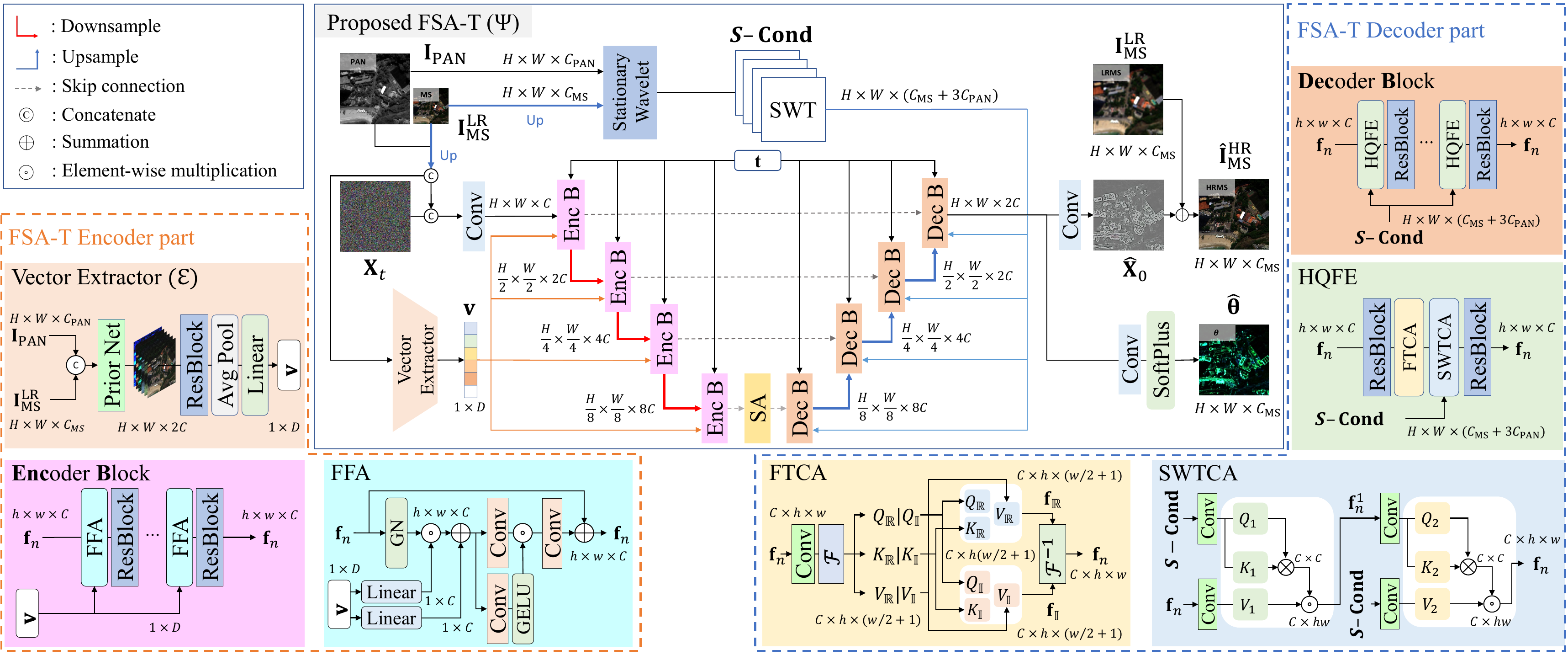}}
    \vspace{-0.6cm}
    \caption{Architecture of our proposed teacher model with frequency-selective attention, denoted as FSA-T, for PAN-sharpening. The FSA-T is designed to fully utilize frequency information for details enhancement from PAN and LRMS inputs (more details in \textit{Supplemental.})}
    \label{Method/FSA_T}
    \vspace{-0.4cm}
\end{figure*}

\subsection{Diffusion Process of Teacher Network (FSA-T)}
\label{sec:32}
Diffusion models operate by defining a forward process that gradually adds noise to data until it becomes indistinguishable from random noise, and a reverse process that learns to denoise the data, reconstructing the original information. We leverage this framework into a frequency-selective attention teacher network (FSA-T) as our teacher network, to generate frequency enhanced HRMS images with their pixel-wise uncertainty information.

\noindent\textbf{Forward process.}\quad In the forward diffusion process, noise is progressively added to the residual image, $\mathbf{X}_{0}$, until it converges to a Gaussian distribution, $\mathbf{X}_{T} \sim \mathcal{N}(\mathbf{0}, \mathbf{I})$, after $T$ timesteps. At each timestep $t \in \{1, \cdots, T\}$, a noisy version of the residual is obtained as follows \cite{ho2020denoising}:
\begin{equation}
    \mathbf{X}_{t} = \sqrt{\bar{\alpha}_t} \mathbf{X}_0 + \sqrt{1 - \bar{\alpha}_t} \, \bm{\epsilon},
\end{equation}
where $\bm{\epsilon} \sim \mathcal{N}(\mathbf{0}, \mathbf{I})$ is Gaussian noise, and $\bar{\alpha}_{t} = \prod_{s=1}^{t} (1 - \beta_{s})$ controls the noise level at step $t$.

\noindent\textbf{Reverse process.}\quad 
Following the DDPM \cite{ho2020denoising}, the reverse process aims to recover $\mathbf{X}_0$ from $\mathbf{X}_T$ by predicting a denoised image at each step. To perform denoising at each timestep, the denoising network (FSA-T), denoted as $\Psi$, is designed to predict the original $\mathbf{X}_0$ rather than the added noise $\bm{\epsilon}$, as we found this empirically to be beneficial for PAN-sharpening. Additionally, $\Psi$ simultaneously produces an uncertainty map $\hat{\bm{\theta}}$ for each prediction, providing pixel-wise confidence information as:
\begin{equation}
    \left[\widehat{\mathbf{X}}_0 \mid \hat{\bm{\theta}}\right] = \Psi\left(\left[\mathbf{X}_{t}\mid\mathbf{I}_\text{PAN}\mid\mathbf{I}_\text{MS}^\text{LR}\right];\;\mathbf{v},\,\mathcal{S}\text{-}\mathsf{Cond};\;t\right),
    \vspace{-0.2cm}
\end{equation}
where $\left[\;\cdot\mid\cdot\;\right]$ indicates channel-wise concatenation, $\widehat{\mathbf{X}}_0$ is a predicted residual, $\hat{\bm{\theta}}$ is a predicted uncertainty map, $\mathbf{v}$ and $\mathcal{S}\text{-}\mathsf{Cond}$ represent a learning compact representation (vector) and Stationary Wavelet Transform (SWT)-based conditioning terms, respectively. This formulation allows the FSA-T to iteratively refine the noisy input, guided by the conditioned information ($\mathbf{v}$, $\mathcal{S}\text{-}\mathsf{Cond}$), toward an accurate reconstruction of the $\mathbf{X}_0$. $\hat{\bm{\theta}}$ captures pixel-wise prediction confidence, which guides the FSA-S to more rigorously learn the regions with high uncertainty. To ensure positivity, we apply the $\mathsf{SoftPlus}$ \cite{zheng2015improving} activation to $\hat{\bm{\theta}}$.

\noindent\textbf{Uncertainty-driven diffusion loss.}\quad Inspired by \cite{ning2021uncertainty, seitzer2022pitfalls}, we employ the following uncertainty-driven diffusion loss function, $\mathcal{L}_\text{U-Diff}$, to optimize the teacher network $\Psi$:
\begin{equation}
    \mathcal{L}_\text{U-Diff} = \left\lVert \frac{1}{2\hat{\bm{\theta}}} \odot \left| \widehat{\mathbf{X}}_0 - \mathbf{X}_0 \right|+ \frac{1}{2}\log \hat{\bm{\theta}} \right\rVert_1,
    \label{eq:U-diff}
\end{equation}
where $\hat{\bm{\theta}}$ serves as the estimated variance term and is regarded as the uncertainty map in our framework. This objective function encourages the model to minimize prediction errors in low-confidence regions, thus yielding a more accurate and spatially robust output.

\subsection{Details of Teacher Network Architecture}
\label{sec:33}
Our teacher network, FSA-T, is designed to enhance frequency details for PAN-sharpening tasks by fully utilizing complementary information from PAN and LRMS inputs. The FSA-T adopts an encoder-decoder (U-Net \cite{ronneberger2015u}) structure, where each stage is specifically tailored to capture and refine critical spatial and frequency features. The encoder of FSA-T employs Feed Forward Attention (FFA) blocks to extract compact and informative representations from the PAN and LRMS inputs. This attention mechanism allows the encoder to focus on essential features, preparing the data for effective frequency enhancement in subsequent layers. 

The decoder of FSA-T integrates High-Quality Frequency Enhancement (HQFE) blocks at each decoding level to further refine frequency components. Each HQFE block comprises two key modules: Fourier Transform Channel Attention (FTCA) that emphasizes important frequency components via Fourier Transform based attention, and Stationary Wavelet Transform Cross Attention (SWTCA) that merges high-frequency information from PAN and low-frequency content from LRMS, both of which are decomposed by SWT \cite{nason1995stationary}. By incorporating these specialize components, FSA-T effectively produces $\widehat{\mathbf{X}}_{0}$ with enhanced spatial fidelity, establishing an effective uncertainty-aware knowledge distillation (U-Know) strategy.

\noindent\textbf{Feed Forward Attention (FFA).}\quad 
The FFA block in Fig.~\ref{Method/FSA_T} is designed to leverage compact information \cite{xia2023diffir} from the PAN and LRMS input to establish a rough structural representation during encoding. A compact vector representation $\mathbf{v} \in \mathbb{R}^{1 \times D}$ is obtained by a vector extractor ($\mathcal{E}$) as:
\begin{equation}
    \mathbf{v} = \mathcal{E}\left(\mathbf{I}_{\text{PAN}}, \mathbf{I}_{\text{MS}}^{\text{LR}}\right).
    \label{eq:vector_extract}
\end{equation}
$\mathbf{v}$ is then used as a dynamic modulation \cite{wang2018recovering, karras2019style} parameter within the FFA, enabling $\mathcal{E}$ to incorporate $\mathbf{v}$ and establish an approximate structure during restoration as follows:
\begin{equation}
    \begin{aligned}
    &[\bm{\beta}_n \mid \bm{\gamma}_n] = \mathsf{Linear}(\mathbf{v}),\\
    &\mathbf{f}_{n}^{\prime} = \bm{\gamma}_n \odot \mathsf{GN}(\mathbf{f}_{n}) + \bm{\beta}_n,\\
    &\mathbf{f}_{n} \leftarrow  \mathsf{Conv}(\mathbf{f}_{n}^{\prime}) \odot \mathsf{GELU}(\mathsf{Conv}(\mathbf{f}_{n}^{\prime})) + \mathbf{f}_{n},
    \end{aligned}
    \label{eq:feedforward}
\end{equation}
where $\mathbf{f}_{n}\in\mathbb{R}^{C \times h \times w}$ denotes input feature for the $n$-th encoder, $\mathsf{Linear}$ indicates Linear operation, $\odot$ denotes element-wise multiplication, $\mathsf{GN}$ indicates group normalization \cite{wu2018group}, $\mathsf{GELU}$ indicates GELU \cite{hendrycks2016gaussian} activation and $\mathsf{Conv}$ consists of a sequential $1 \times 1$ convolution and a $3 \times 3$ depth-wise convolution layer (Details of $\mathcal{E}$ is in the \textit{Supplemental.}).

\noindent\textbf{Fourier Transform Channel Attention (FTCA).}\quad 
The FTCA block in Fig.~\ref{Method/FSA_T} performs self-attention in frequency domain by leveraging the 2D Fast Fourier Transform (FFT), denoted as $\mathcal{F}$. $\mathcal{F}$ allows for efficient identification and emphasis of critical frequency components. For the $n$-th decoder block in Fig.~\ref{Method/FSA_T}, the input feature $\mathbf{f}_{n}\in\mathbb{R}^{C \times h \times w}$ of $C$ channels is transformed into query ($\mathbf{Q}$), key ($\mathbf{K}$), and value ($\mathbf{V}$) representations. The 2D FFT $\mathcal{F}$ is then applied independently to each channel of $\mathbf{Q}$, $\mathbf{K}$, and $\mathbf{V}$, mapping them transform from time domain to frequency domain as:
\begin{equation}
    \left[ \mathbf{Q} \mid \mathbf{K} \mid \mathbf{V} \right] = \mathcal{F}\left(\mathsf{Conv}\left(\mathbf{f}_{n}\right)\right),
\end{equation}
where $\mathbf{Q}$, $\mathbf{K}$ and $\mathbf{V}$ reside in $\mathbb{C}^{C \times h \times (\frac{w}{2} + 1)}$ due to the Hermitian symmetry property \cite{celeghini2021hermite}. Note that each frequency domain representation of $\mathbf{Q}$, $\mathbf{K}$, and $\mathbf{V}$ includes real and imaginary parts. So, we decompose $\mathbf{Q}$, $\mathbf{K}$, and $\mathbf{V}$ into real and imaginary parts as:
\begin{equation}
    \left[ \mathbf{Q}_{\mathbb{R}} \mid \mathbf{K}_{\mathbb{R}} \mid \mathbf{V}_{\mathbb{R}} \right] + j \left[ \mathbf{Q}_{\mathbb{I}} \mid \mathbf{K}_{\mathbb{I}} \mid \mathbf{V}_{\mathbb{I}} \right] = \left[ \mathbf{Q} \mid \mathbf{K} \mid \mathbf{V} \right],
\end{equation}
where $j=\sqrt{-1}$, and $\mathbf{Q}_{\mathbb{R}}$, $\mathbf{K}_{\mathbb{R}}$, $\mathbf{V}_{\mathbb{R}}$, $\mathbf{Q}_{\mathbb{I}}$, $\mathbf{K}_{\mathbb{I}}$, $\mathbf{V}_{\mathbb{I}}$ are in $\mathbb{R}^{C \times h \times (\frac{w}{2} + 1)}$. Self-attention operations are then performed separately on the real and imaginary parts as follows:

\begin{equation}
    \mathbf{f}_{l} = \mathsf{SoftMax}\left(\mathbf{Q}_{l} \mathbf{K}_{l}^{\mathsf{T}}/\sqrt{C}\right) \mathbf{V}_{l},
\end{equation}
where $l \in \{\mathbb{R}, \mathbb{I}\}$ indicates the real or imaginary feature, respectively. Finally, the inverse Fourier Transform $\mathcal{F}^{-1}$ is applied to $\mathbf{f}_{\mathbb{R}}$ and $\mathbf{f}_{\mathbb{I}}$ to map them back to the time domain:
\begin{equation}
    \mathbf{f}_{n} \leftarrow \mathcal{F}^{-1}\left(\mathbf{f}_{\mathbb{R}} + j \mathbf{f}_{\mathbb{I}}\right).
\end{equation}
This process ensures that refined frequency information be selectively captured and utilized.

\noindent\textbf{Stationary Wavelet Transform Cross Attention (SWTCA).}\quad 
The SWTCA block is designed to leverege Wavelet components of $\mathbf{I}_\text{PAN}$ and $\mathbf{I}_\text{MS}^\text{LR}$ motivated by \cite{shang2024resdiff, phung2023wavelet, zhou1998wavelet}. As shown in Fig.~\ref{Method/FSA_T}, it enhances the features outputted by the FTCA block, utilizing the Stationary Wavelet Transform (SWT) \cite{nason1995stationary}, denoted as $\mathcal{S}$. The SWTCA block selectively incorporates relevant frequency components from $\mathbf{I}_\text{PAN}$ and $\mathbf{I}_\text{MS}^\text{LR}$ through a two-stage cross-attention process. The SWT-based conditioning terms ($\mathcal{S}\text{-}\mathsf{Cond}$) are constructed by decomposing $\mathbf{I}_\text{PAN}$ and $\mathbf{I}_\text{MS}^\text{LR}$ using $\mathcal{S}$ into four components:
\begin{equation}
    \begin{aligned}
    \left[\mathbf{L}_\text{PAN} \mid \mathbf{H}_\text{PAN} \mid \mathbf{V}_\text{PAN} \mid \mathbf{D}_\text{PAN}\right] &= \mathcal{S}(\mathbf{I}_\text{PAN}), \\
    \left[\mathbf{L}_\text{MS}^\text{LR} \mid \mathbf{H}_\text{MS}^\text{LR} \mid \mathbf{V}_\text{MS}^\text{LR} \mid \mathbf{D}_\text{MS}^\text{LR}\right] &= \mathcal{S}(\mathbf{I}_\text{MS}^\text{LR}),
    \end{aligned}
\end{equation}

where $\mathbf{L}$ represent the low-frequency approximation components, while $\mathbf{H}$, $\mathbf{V}$, $\mathbf{D}$ denote the high-frequency horizontal, vertical, and diagonal details, respectively. Since $\mathbf{I}_\text{PAN}$ provides high spatial resolution but lacks spectral information, only its high-frequency components $\mathbf{H}_\text{PAN}$, $\mathbf{V}_\text{PAN}$, and $\mathbf{D}_\text{PAN}$ are used. Conversely, $\mathbf{L}_\text{MS}^\text{LR}$ from $\mathbf{I}_\text{MS}^\text{LR}$ contains richer spectral information. These components are concatenated channel-wise to form $\mathcal{S}\text{-}\mathsf{Cond}$ as:
\begin{equation}
    \mathcal{S}\text{-}\mathsf{Cond} = \left[\mathbf{L}_\text{MS}^\text{LR} \mid \mathbf{H}_\text{PAN} \mid \mathbf{V}_\text{PAN} \mid \mathbf{D}_\text{PAN}\right].
\end{equation}
To inject $\mathcal{S}\text{-}\mathsf{Cond}$ into the feature $\mathbf{f}_{n}$ obtained from FTCA at the $n$-th decoder block, we employ a two-stage cross-attention process. In the first stage, $\mathbf{Q}_{1}$ and $\mathbf{K}_{1}$ are derived from $\mathcal{S}\text{-}\mathsf{Cond}$, directing attention toward critical frequency components, while $\mathbf{V}_{1}$ is derived from $\mathbf{f}_{n}$. The resulting features are reshaped to $C \times hw$ for channel attention \cite{zamir2022restormer}, producing an intermediate feature $\mathbf{f}_{n}^{1} \in \mathbb{R}^{C \times h \times w}$ as:

\begin{equation}
    \begin{aligned}
    &\left[\mathbf{Q}_{1} \mid \mathbf{K}_{1}\right] = \mathsf{Conv} (\mathcal{S}\text{-}\mathsf{Cond}),\;\mathbf{V}_{1} = \mathsf{Conv}(\mathbf{f}_{n}), \\
    &\mathbf{f}_{n}^{1} = \mathsf{SoftMax}\left(\mathbf{Q}_{1} \mathbf{K}_{1}^{\mathsf{T}}/{\sqrt{C}}\right) \mathbf{V}_{1},
    \end{aligned}
\end{equation}
where $\mathbf{Q}_{1}$, $\mathbf{K}_{1}$, $\mathbf{V}_{1}$ are in $\mathbb{R}^{C \times hw}$. In the second stage, $\mathbf{Q}_{2}$ is derived from the intermediate feature $\mathbf{f}_{n}^{1}$, while $\mathbf{K}_{2}$ and $\mathbf{V}_{2}$ come from $\mathcal{S}\text{-}\mathsf{Cond}$, enabling the model to refine the correlation between the reconstructed feature $\mathbf{f}_{n}$ and significant frequency components in $\mathcal{S}\text{-}\mathsf{Cond}$. This stage yields the final feature $\mathbf{f}_{n} \in \mathbb{R}^{C \times h \times w}$ after reshaping:

\begin{equation}
    \begin{aligned}
    &\mathbf{Q}_{2} = \mathsf{Conv}(\mathbf{f}_{n}^{1}), \; \left[\mathbf{K}_{2} \mid \mathbf{V}_{2}\right] = \mathsf{Conv}(\mathcal{S}\text{-}\mathsf{Cond}), \\
    &\mathbf{f}_{n} \leftarrow \mathsf{SoftMax}\left(\mathbf{Q}_{2} \mathbf{K}_{2}^{\mathsf{T}}/{\sqrt{C}}\right) \mathbf{V}_{2}.
    \end{aligned}
\end{equation}
Through these two stages of cross-attention, the final feature \(\mathbf{f}_{n}\) effectively incorporates essential frequency components from both PAN and MS inputs, ensuring enhanced frequency representation and high-quality HRMS output.

\subsection{Uncertainty-Aware Knowledge Distillation}
\label{sec:34}
To address the substantial computational demands of the teacher (FSA-T) $\Psi$, we propose an novel uncertainty-aware knowledge distillation (U-Know) strategy.

\noindent\textbf{Student network (FSA-S) $\psi$.}\quad The student model FSA-S is designed with the same number of encoder and decoder blocks as FSA-T. Each block consists solely of Resblocks, and unlike the teacher, the student operates without any additional conditional inputs. As shown in Fig.~\ref{Method/kd_model}, the lightweight student model $\psi$ predicts the denoised image $\widetilde{\mathbf{X}}_{0}$ from the input $\mathbf{X}_{t}$ at timestep $t$ as:
\begin{equation}
    \widetilde{\mathbf{X}}_{0} = \psi\left(\left[\mathbf{X}_{t}\mid\mathbf{I}_\text{PAN}\mid\mathbf{I}_\text{MS}^\text{LR}\right]; t\right).
\end{equation}

\noindent\textbf{U-Know loss function.}\quad FSA-T $\Psi$ generates an uncertainty map $\hat{\bm{\theta}}$ along with the output $\widehat{\mathbf{X}}_{0}$ during training. These two outputs are frozen and used in the U-Know loss function, $\mathcal{L}_{\text{U-Know}}$, to train the FSA-S $\psi$. The total training objective combines a hard loss ($\mathcal{L}_\text{hard}$), a soft loss ($\mathcal{L}_\text{soft}$), and a feature loss ($\mathcal{L}_\text{feat}$) to distill frequency-rich features and enhance regions where $\Psi$ exhibits spatial weaknesses points. The total loss is defined as:
\begin{equation}
    \mathcal{L}_\text{U-Know} = \mathcal{L}_\text{hard} + \lambda_{s}\mathcal{L}_\text{soft} + \lambda_{f}\mathcal{L}_\text{feat},
\end{equation}
where $\lambda_{s}$ and $\lambda_{f}$ control the contributions of $\mathcal{L}_\text{soft}$ and $\mathcal{L}_\text{feat}$, respectively.  $\mathcal{L}_{\text{hard}}$ measures the difference between $\psi$ output $\widetilde{\mathbf{X}}_{0}$ and the ground truth $\mathbf{X}_{0}$, weighted by $\Psi$’s uncertainty map $\hat{\bm{\theta}}$ to emphasize challenging regions:
\begin{equation}
    \mathcal{L}_\text{hard} = \left\lVert \left(\tau + \hat{\bm{\theta}}\right)\odot\left|\widetilde{\mathbf{X}}_{0} - \mathbf{X}_{0}\right| \right\rVert_1,
\end{equation}
where $\tau$ is a hyper-parameter. $\mathcal{L}_{\text{soft}}$ computes the difference between $\psi$’s output $\widetilde{\mathbf{X}}_{0}$ and $\Psi$’s output $\widehat{\mathbf{X}}_{0}$, weighted by the $\hat{\bm{\theta}}$ to help $\psi$ better approximate the $\Psi$’s knowledge:
\begin{equation}
    \mathcal{L}_\text{soft} = \left\lVert \left(\tau - \hat{\bm{\theta}}\right)\odot\left|\widetilde{\mathbf{X}}_{0} - \widehat{\mathbf{X}}_{0}\right| \right\rVert_1.
\end{equation}
In the regions where $\hat{\bm{\theta}}$ values are high, indicating that $\Psi$ struggles with the task, we apply a weight of $(\tau + \hat{\bm{\theta}})$ and use $\mathbf{X}_0$ as the hard ground truth to guide $\psi$ toward the true target. Conversely, in the areas where $\hat{\bm{\theta}}$ values are low, indicating that $\Psi$ performs well, we apply a weight of $(\tau - \hat{\bm{\theta}})$ and use $\Psi$’s output $\widehat{\mathbf{X}}_0$ as the soft ground truth, allowing $\psi$ to approximate $\Psi$’s knowledge in these regions. These dual loss formulations enable $\Psi$ to focus on challenging areas while effectively leveraging the $\Psi$’s strengths. $\mathcal{L}_{\text{feat}}$ ensures alignment between intermediate features of $\Psi$ and $\psi$ as \cite{barron2019general}:
\begin{equation}
    \ell_\text{feat}^{i} = \sqrt{\left\lVert \mathbf{f}_{i} - \mathbf{s}_{i} \right\rVert_1^2 + \gamma},\quad\mathcal{L}_\text{feat} = \sum\nolimits_{i}{\alpha_i} \ell_\text{feat}^{i},
\label{eq:L_feat}
\end{equation}
where $\ell_\text{feat}^{i}$ is a feature loss for the $i$-th feature, $\alpha_i$ is a weight for the $i$-th feature, and $\mathbf{f}_{i}$ and $\mathbf{s}_{i}$ denote the $i$-th intermediate feature maps of $\Psi$ and $\psi$ respectively. The small constant $\gamma$ ensures smooth convergence during optimization.
\section{Experiment}
\subsection{Datasets, Metrics and Training Details}

\begin{table}[htbp]
\scriptsize
\centering

\resizebox{0.99\columnwidth}{!}{
\def\arraystretch{1.2}
\begin{tabular}{c|c|c|c|c}
\Xhline{2\arrayrulewidth}
\multicolumn{2}{c|}{\textbf{Satellite}} & \textbf{WorldView-3} & \textbf{QuickBird} & \textbf{GaoFen-2}\\
\hline
\multicolumn{2}{c|}{Number of Band} & 8 & 4 & 4\\
\hline
\multirow{2}{*}{Spatial Resolution (m)} & PAN & 0.3 & 0.6 & 0.8\\
\cline{2-5}
& LRMS & 1.2 & 2.4 & 3.2\\
\hline
\multicolumn{2}{c|}{Radiometric Resolution (bit)} & 11 & 11 & 10\\
\hline
\multicolumn{2}{c|}{Number of (Train / Test) Images} & 9,714 / 20  & 17,139 / 20 & 19,809 / 20\\
\hline
\multirow{2}{*}{Patch Size} & PAN  & 64×64×1 & 64×64×1 & 64×64×1\\
\cline{2-5}
& LRMS & 16×16×8 & 16×16×4 & 16×16×4\\
\Xhline{2\arrayrulewidth}
\end{tabular}}
\caption{Detailed information of Worldview-3, QuickBird, and GaoFen-2 datasets.}
\vspace{-0.3cm}
\label{tab:dataset}
\end{table}

\begin{table*}[tbp]
\scriptsize
\centering
\resizebox{0.99\textwidth}{!}{ 
\def\arraystretch{1.2}
\begin{tabular}{c|l|l|cccccc|ccc}
\Xhline{2\arrayrulewidth}
\multicolumn{3}{c|}{\textbf{GF2} Dataset} & \multicolumn{6}{c|}{Reduced-Resolution}&\multicolumn{3}{c}{Full-Resolution}\\
\hline
Types & Methods & Publications & PSNR$\uparrow$ & SSIM$\uparrow$ & SAM$\downarrow$ & ERGAS$\downarrow$ & SCC$\uparrow$ & Q4$\uparrow$ & $D_{\lambda} \downarrow$ & $D_{s}\downarrow$ & HQNR$\uparrow$\\
\hline
\multirow{7}{*}{Non-DMs} & PanNet \cite{yang2017pannet} & ICCV 2017 & 39.197 & 0.959 & 1.050 & 1.038 & 0.975 & 0.963 & 0.020 & 0.052 & 0.929 \\
& MSDCNN \cite{yuan2018multiscale} & JSTARS 2018 & 40.730 & 0.971 & 0.946 & 0.862 & 0.983 & 0.972 & 0.026 & 0.079 & 0.898 \\
& FusionNet \cite{wu2021dynamic} & ICCV 2021 & 39.866 & 0.966 & 0.971 & 0.960 & 0.980 & 0.967 & 0.034 & 0.105 & 0.865 \\
& LAGConv \cite{jin2022lagconv} & AAAI 2022 & 41.147 & 0.974 & 0.886 & 0.816 & 0.985 & 0.974 & 0.030 & 0.078 & 0.895 \\
& S2DBPN \cite{zhang2023spatial} & TGRS 2023 & 42.686 & 0.980 & 0.772 & 0.686 & 0.990 & 0.981 & 0.020 & 0.046 & 0.935 \\
& DCPNet \cite{zhang2024dcpnet} & TGRS 2024 & 42.312 & 0.979 & 0.806 & 0.724 & 0.988 & 0.980 & 0.024 & {\color{red}{\textbf{0.024}}} & {\color{red}{\textbf{0.953}}} \\
& CANConv \cite{duan2024content} & CVPR 2024 & 43.166 & 0.982 & 0.722 & 0.653 & 0.991 & 0.983 & 0.019 & 0.063 & 0.919 \\
\hline
\multirow{4}{*}{{\makecell{Diffusion\\Models}}} & PanDiff \cite{meng2023pandiff} & TGRS 2023 & 42.827 & 0.980 & 0.767 & 0.674 & 0.990 & 0.981 & 0.020 & 0.045 & 0.936 \\
& TMDiff \cite{xing2024empower} & TGRS 2024 & 41.896 & 0.977 & 0.764 & 0.754 & 0.988 & 0.979 & 0.029 & 0.030 & 0.942 \\
& \textbf{FSA-S} & \textbf{Ours}& {\color{blue}{\underline{44.585}}} & {\color{blue}{\underline{0.986}}} & {\color{blue}{\underline{0.624}}} & {\color{blue}{\underline{0.548}}} & {\color{blue}{\underline{0.993}}} & {\color{blue}{\underline{0.987}}} & {\color{blue}{\underline{0.018}}} & 0.037 & {\color{blue}{\underline{0.944}}} \\
& \textbf{FSA-T} & \textbf{Ours} & {\color{red}{\textbf{44.757}}} & {\color{red}{\textbf{0.988}}} & {\color{red}{\textbf{0.603}}} & {\color{red}{\textbf{0.537}}} & {\color{red}{\textbf{0.994}}} & {\color{red}{\textbf{0.988}}} & {\color{red}{\textbf{0.017}}} & {\color{blue}{\underline{0.029}}} & {\color{red}{\textbf{0.953}}} \\
\Xhline{2\arrayrulewidth}
\end{tabular}}
\vspace{-0.25cm}
\caption{Comparison of different models on the GaoFen-2 (GF2) dataset. {\color{blue}\underline{Blue}} indicates the second-best performance, while \textbf{\color{red}red} highlights the best-performing model. Standard deviations of the metrics across test samples are provided in the \textit{Supplemental}.}
\label{tab:gf2}
\end{table*}

\begin{table*}[tbp]
\scriptsize
\centering

\resizebox{0.99\textwidth}{!}{
\def\arraystretch{1.2}
\begin{tabular}{c|l|cccccc|cccccc}
\Xhline{2\arrayrulewidth}
\multirow{2}{*}{Types} & \multirow{2}{*}{Methods} & \multicolumn{6}{c|}{\textbf{WV3} Dataset (Reduced-Resolution)}& \multicolumn{6}{c}{\textbf{QB} Dataset (Reduced-Resolution)} \\
\cline{3-14}
& & PSNR$\uparrow$ & SSIM$\uparrow$ & SAM$\downarrow$ & ERGAS$\downarrow$ & SCC$\uparrow$ & Q8$\uparrow$ & PSNR$\uparrow$ & SSIM$\uparrow$ & SAM$\downarrow$ & ERGAS$\downarrow$ & SCC$\uparrow$ & Q4$\uparrow$\\
\hline
\multirow{7}{*}{Non-DMs} & PanNet \cite{yang2017pannet} & 36.148 & 0.966 & 3.402 & 2.538 & 0.979 & 0.913 & 35.563 & 0.939 & 5.273 & 4.856 & 0.966 & 0.911 \\
& MSDCNN \cite{yuan2018multiscale} & 36.329 & 0.967 & 3.300 & 2.489 & 0.979 & 0.914 & 37.040 & 0.954 & 4.828 & 4.074 & 0.977 & 0.925 \\
& FusionNet \cite{wu2021dynamic} & 36.569 & 0.968 & 3.188 & 2.428 & 0.981 & 0.916 & 36.821 & 0.952 & 4.892 & 4.183 & 0.975 & 0.923 \\
& LAGConv \cite{jin2022lagconv} & 36.732 & 0.970 & 3.153 & 2.380 & 0.981 & 0.916 & 37.565 & 0.958 & 4.682 & 3.845 & 0.980 & 0.930 \\
& S2DBPN \cite{zhang2023spatial} & 37.216 & 0.972 & 3.019 & 2.245 & 0.985 & 0.917 & 37.314 & 0.956 & 4.849 & 3.956 & 0.980 & 0.928 \\
& DCPNet \cite{zhang2024dcpnet} & 37.009 & 0.972 & 3.083 & 2.301 & 0.984 & 0.915 & 38.079 & {\color{blue}{\underline{0.963}}} & 4.420 & 3.618 & 0.983 & 0.935 \\
& CANConv \cite{duan2024content} & 37.441 & {\color{blue}{\underline{0.973}}} & 2.927 & 2.163 & 0.985 & 0.918 & 37.795 & 0.960 & 4.554 & 3.740 & 0.982 & {\color{blue}{\underline{0.935}}} \\
\hline
\multirow{4}{*}{{\makecell{Diffusion\\Models}}} & PanDiff \cite{meng2023pandiff} & 37.029 & 0.971 & 3.058 & 2.276 & 0.984 & 0.913 & 37.842 & 0.959 & 4.611 & 3.723 & 0.982 & {\color{blue}{\underline{0.935}}} \\
& TMDiff \cite{xing2024empower} & 37.477 & 0.973 & 2.885 & 2.151 & 0.986 & 0.915 & 37.642 & 0.958 & 4.627 & 3.804 & 0.981 & 0.930 \\
& \textbf{Our FSA-S}   & {\color{red}{\textbf{37.930}}}
& {\color{red}{\textbf{0.976}}} & {\color{red}{\textbf{2.797}}} & {\color{red}{\textbf{2.046}}} & {\color{red}{\textbf{0.988}}} & {\color{red}{\textbf{0.922}}} & {\color{red}{\textbf{38.361}}}
& {\color{red}{\textbf{0.964}}} & {\color{red}{\textbf{4.337}}} & {\color{red}{\textbf{3.500}}} & {\color{blue}{\underline{0.984}}} & {\color{red}{\textbf{0.938}}} \\
& \textbf{Our FSA-T}   & {\color{blue}{\underline{37.894}}} & {\color{red}{\textbf{0.976}}} & {\color{blue}{\underline{2.801}}} & {\color{blue}{\underline{2.055}}} & {\color{blue}{\underline{0.987}}} & {\color{blue}{\underline{0.921}}} & {\color{blue}{\underline{38.343}}} & {\color{red}{\textbf{0.964}}} & {\color{blue}{\underline{4.349}}} & {\color{blue}{\underline{3.502}}} & {\color{red}{\textbf{0.985}}} & {\color{red}{\textbf{0.938}}} \\
\Xhline{2\arrayrulewidth}
\end{tabular}}
\vspace{-0.2cm}
\caption{Performance comparison of different models on WV3 and QB datasets. \textbf{\color{red}Red} and {\color{blue}\underline{blue}} highlights the best- and 2nd-best performing models. The results for the full-resolution datasets and detailed results with standard deviations are provided in the \textit{Supplemental}.}
\vspace{-0.3cm}
\label{tab:main}
\end{table*}

\begin{figure*}[!tbp]
     \centering{\includegraphics[width=\textwidth]{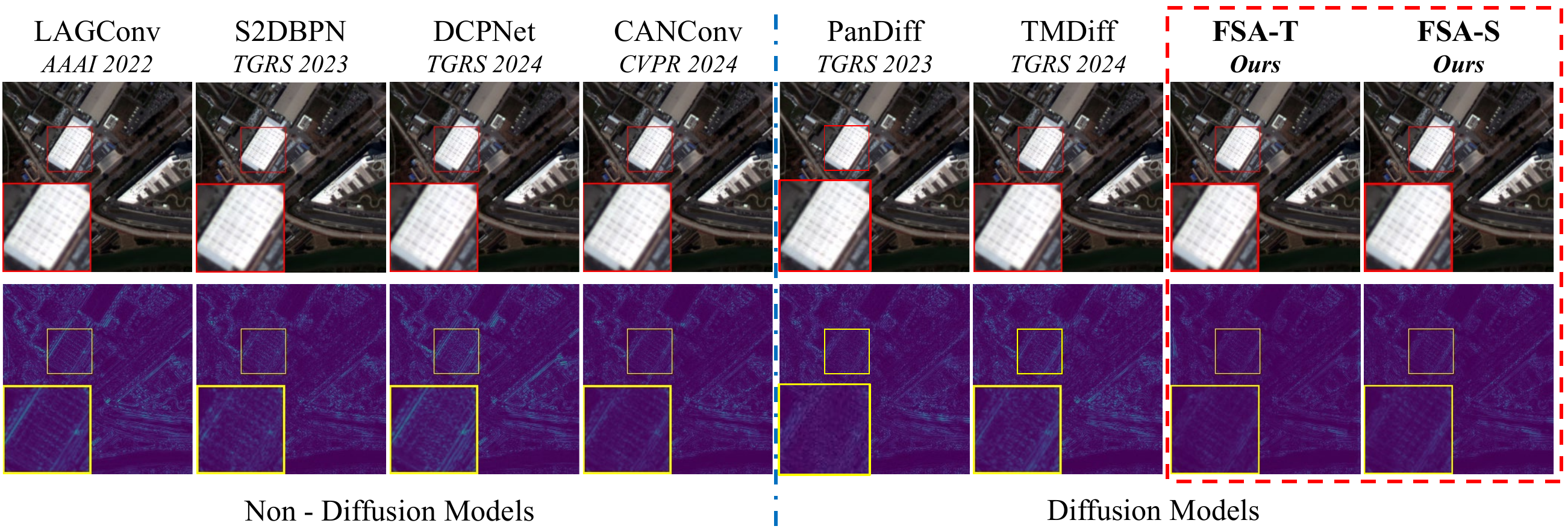}}
     \vspace{-0.6cm}
     \caption{Visual comparison of PAN-sharpening results on the reduced-resolution GF2 dataset. The first row shows RGB of outputs $\textbf{I}^\text{HR}_\text{MS}$, and the second row displays Error Map, the difference between output $\hat{\textbf{I}}^\text{HR}_\text{MS}$ and ground truth $\textbf{I}^\text{HR}_\text{MS}$. Both FSA-T and FSA-S achieve more detailed results compared to state-of-the-art models.}
     \label{Experiment/qualitative_reduced_gf2}
     \vspace{-0.2cm}
\end{figure*}

Table~\ref{tab:dataset} summarizes the characteristics of three datasets from PanCollection \cite{deng2022machine} for our experiments: WorldView-3 (WV3), QuickBird (QB), and GaoFen-2 (GF2). Data augmentation was performed using random vertical and horizontal flips. All experiments were conducted with a batch size of 32 for 300K iterations per dataset on an NVIDIA GeForce RTX 3090 GPU. The initial learning rate was set to $10^{-4}$ with a decay factor of 0.5 applied every 10K iterations, using the AdamW optimizer \cite{diederik2014adam} with $\beta_1 = 0.9$, $\beta_2 = 0.999$, and a weight decay of $10^{-4}$. For the diffusion process of our FSA-T and FSA-S, the total diffusion time step $T$ was set to 500, with the DDIM \cite{song2020denoising} sampling strategy used to generate HRMS images in 25 steps. We set the hyper-parameters as $\lambda_s = 0.1$, $\lambda_f = 0.001$, $\tau = 1$, and $\alpha_{i} = 1$, which were empirically found to yield excellent performance. To evaluate the model performance, we employed widely accepted metrics that capture various aspects of image quality and similarity. For the reduced-resolution datasets of WV3, QB, and GF2, we used six standard metrics: PSNR \cite{wang2004image}, SSIM \cite{wang2004image}, SAM \cite{yuhas1992discrimination}, ERGAS \cite{wald2000quality}, SCC \cite{garzelli2009hypercomplex}, and Q4/Q8 \cite{vivone2014critical}. For the full-resolution datasets of WV3, QB, and GF2, we used HQNR \cite{vivone2020new} with $D_{\lambda}$, and $D_S$ to comprehensively assess the results.

\subsection{Experimental Results}
To ensure a thorough and equitable evaluation, we used the official codes for each compared method whenever available. For methods without official codes, we re-implemented them based on the papers \cite{meng2023pandiff, xing2024empower} to maintain fairness in comparison. 

\noindent\textbf{Quantitative analysis.}\quad As shown in Table~\ref{tab:gf2} and Table~\ref{tab:main}, our FSA-T achieves superior performance on the WV3, QB, and GF2 datasets compared to all other methods. The GF2 dataset contains relatively easier (less complex) images, so the teacher model can fully leverage its high-capacity features. In contrast, for the more challenging (more complex) WV3 and QB datasets, our FSA-S surpasses FSA-T across several metrics. This improvement represents the effectiveness of our uncertainty-aware knowledge distillation (U-Know) strategy, which enables the student model (FSA-S) to address the spatial weaknesses of the teacher model by focusing on uncertain regions. From this, the FSA-S can benefit more generalization from the knowledge distillation with uncertainty maps, leading to better performances than the FSA-T for the test datasets. Consequently, our FSA-S demonstrates an enhanced adaptability and robustness on complex datasets, where precise handling of spatial details (high frequency components) is essential. 

\noindent\textbf{Qualitative analysis.}\quad
Fig. \ref{Experiment/qualitative_reduced_gf2} shows the output HRMS images $\hat{\textbf{I}}^{\text{HR}}_{\text{MS}}$  in the first rows and the error maps (MAE images) between $\hat{\textbf{I}}^{\text{HR}}_{\text{MS}}$ and ground truth $\textbf{I}^{\text{HR}}_{\text{MS}}$. As shown, our framework generates the results $\hat{\textbf{I}}^{\text{HR}}_{\text{MS}}$ closer to $\textbf{I}^{\text{HR}}_{\text{MS}}$ in the reduced resolution GF2 data set. Furthermore, Fig. \ref{fig:first} demonstrates that even in the full-resolution comparison where ground truth is unavailable, our framework enables to generate more detailed and stable results compared to all other state-of-the-art approaches. 

\begin{table}[htbp]
\scriptsize
\centering

\resizebox{0.85\columnwidth}{!}{
\def\arraystretch{1.2}
\begin{tabular}{l|cccc}
\Xhline{2\arrayrulewidth}
Methods & Params. (M) & FLOPs (T) & Time (s) & Memory (GB)\\
\hline
PanDiff \cite{meng2023pandiff} & 12.556 & 0.471 & 19.522 & 3.260\\
TMDiff \cite{xing2024empower} & 153.939 & 5.517 & 67.461 & 10.483\\
\textbf{FSA-T} & 25.492 & 1.402 & 25.495 & 5.910\\
\textbf{FSA-S} & {\color{red}{\textbf{9.115}}} & {\color{red}{\textbf{0.346}}} & {\color{red}{\textbf{12.287}}} & {\color{red}{\textbf{2.136}}}\\
\Xhline{2\arrayrulewidth}
\end{tabular}}
\caption{Computational complexity comparison for diffusion-based models. Best values are highlighted in {\color{red}{\textbf{red}}}.}

\label{tab:complex}
\vspace{-0.3cm}
\end{table}

\noindent\textbf{Computational complexity analysis.}\quad Table~\ref{tab:complex} compares the computational cost of diffusion-based models, detailing parameters (Params.), floating-point operations (FLOPs), inference time (Time) and memory usage. This demonstrates the efficiency of our FSA-S over other models. In particular, the TMDiff has significantly higher numbers of parameters and FLOPs. The FSA-T, while computationally intensive than the FSA-S, still provides a balanced trade-off between efficiency and performance.

\subsection{Ablation Studies}
\begin{figure}[htbp]
    \centering{\includegraphics[width=1.0\columnwidth]{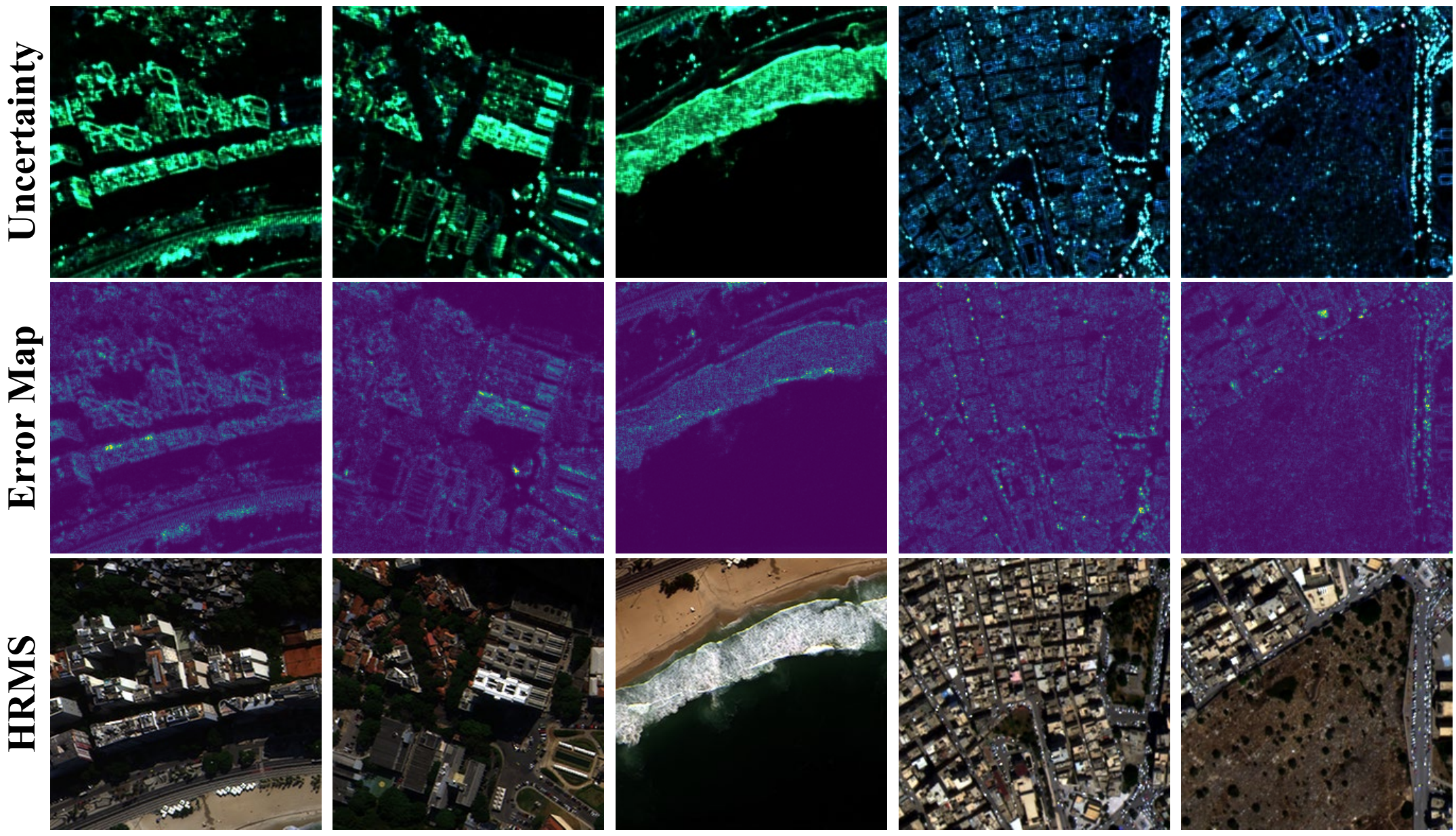}}
    \vspace{-0.6cm}
    \caption{Visualization of uncertainty map $\hat{\bm\theta}$, Error Map, and ground truth $\textbf{I}_\text{MS}^\text{HR}$ of reduced WV3.}
    \label{fig:uncertainty}
    \vspace{-0.2cm}
\end{figure}
\noindent\textbf{Visualization of uncertainty map $\hat{\bm\theta}$ of $\Psi$.}\quad
The uncertainty map $\hat{\bm\theta}$ learned through Eq.~\ref{eq:U-diff} shows that the areas with low uncertainty appear dark, while the areas with high uncertainty appear bright. Fig.~\ref{fig:uncertainty} visualizes the uncertainty maps $\hat{\bm\theta}$, error maps and ground truth $\textbf{I}^{\text{HR}}_{\text{MS}}$ for the reduced WV3 dataset. As shown in Fig.~\ref{fig:uncertainty}, it can be noted that the high-uncertainty regions mainly correspond to high-frequency areas such as challenging object edges. The learned uncertainty maps tend to resemble the error maps.

\begin{table}[htbp]
    \scriptsize
    \centering
    
    \resizebox{0.99\columnwidth}{!}{
    \def\arraystretch{1.2}
    \begin{tabular} {c|c|c|c|c|c}
    \Xhline{2\arrayrulewidth}
    Encoder & Decoder & \multicolumn{4}{c}{\textbf{GF2} Dataset (Reduced-Resolution)} \\
    \hline
    FFA & HQFE & SAM$\downarrow$ & ERGAS$\downarrow$ & SCC$\uparrow$ & Q4$\uparrow$\\
    \hline
    & & 0.654 $\pm$ 0.112 & 0.661 $\pm$ 0.076 & 0.992 $\pm$ 0.001 & 0.986 $\pm$ 0.007 \\
    \checkmark & & $0.654 \pm 0.112$ & $0.636 \pm 0.077$ & $0.993 \pm 0.002$ & $0.986 \pm 0.007$ \\
     & \checkmark & $0.617 \pm 0.117$ & $0.556 \pm 0.104$ & $0.993 \pm 0.002$ & $0.987 \pm 0.007$ \\
    \checkmark & \checkmark & {\color{red}{\textbf{0.603 $\pm$ 0.102}}} & {\color{red}{\textbf{0.537 $\pm$ 0.077}}} & {\color{red}{\textbf{0.994 $\pm$ 0.001}}} & {\color{red}{\textbf{0.988 $\pm$ 0.006}}} \\
    \Xhline{2\arrayrulewidth}
    \end{tabular}}
    \caption{Comparison of Results with and without FFA and HQFE Blocks in FSA-T. Best values are highlighted in {\color{red}{\textbf{red}}}.}
  \label{tab:Effect of FFA, HQFE blocks}
\end{table}

\noindent\textbf{Effect of FFA, HQFE blocks in FSA-T.}\quad
Table~\ref{tab:Effect of FFA, HQFE blocks} presents the effectiveness of FFA and HQFE. As noted, there are a significant improvement across all metrics. The performance gains are much greater than when applying FFA to the encoder alone or HQFE to the decoder alone, as simultaneous condition injection creates a synergistic effect. The synergy of using both FFA and HQFE occurs because FFA sets up initial styles or "sketches" for $\hat{\textbf{I}}^\text{HR}_\text{MS}$ images, followed by the decoder that restores frequency details via the HQFE. This step-by-step learning process greatly benefits overall performance.

\begin{table}[htbp]
    \scriptsize
    \centering
    
    \resizebox{0.85\columnwidth}{!}{
    \def\arraystretch{1.2}
    \begin{tabular} {c|c|c|c}
    \Xhline{2\arrayrulewidth}
    \ \multirow{2}{*}{Loss func.} & \multicolumn{3}{c}{\textbf{GF2} Dataset (Full-Resolution)} \\
    \cline{2-4}
     & $\text{D}_{\lambda}\downarrow$ & $\text{D}_s\downarrow$ & $\text{HQNR}\uparrow$ \\
    \hline
    $\mathcal{L}_1$ & 0.026 $\pm$ 0.014 & 0.040 $\pm$ 0.017 & 0.935 $\pm$ 0.020 \\
    $\mathcal{L}_{\text{KD}}$ & 0.025 $\pm$ 0.015 & 0.038 $\pm$ 0.016 & 0.938 $\pm$ 0.021 \\
    $\mathcal{L}_{\text{U-know}}$ & {\color{red}{\textbf{0.018 $\pm$ 0.011}}} & {\color{red}{\textbf{0.037 $\pm$ 0.007}}} & {\color{red}{\textbf{0.944 $\pm$ 0.012}}} \\
    \Xhline{2\arrayrulewidth}
    \end{tabular}}
    \caption{Impact of our U-know loss function design. Best values are highlighted in {\color{red}{\textbf{red}}}.}
    \vspace{-0.2cm}
  \label{tab:impact of uknow}
\end{table}

\noindent\textbf{Impact of the U-Know loss.}\quad
To verify the effectiveness of our uncertainty-aware knowledge distillation loss $\mathcal{L}_\text{U-know}$ (U-know loss), we trained the student model (FSA-S) using three losses: a widely used $L_1$, a conventional knowledge distillation loss $\mathcal{L}_\text{KD}$ which is same from our method $\mathcal{L}_\text{U-know}$ but without considering the uncertainty map, and our $\mathcal{L}_\text{U-know}$. Table~\ref{tab:impact of uknow} shows the performance comparisons for the three losses. As shown, when the FSA-S was trained with $\mathcal{L}_\text{U-know}$, it achieved the best performance across all three metrics. This highlights the effectiveness of $\mathcal{L}_\text{U-know}$ that enables the FSA-S to better leverage the teacher (FSA-T)'s prior knowledge including the uncertainty maps $\hat{\bm{\theta}}$, output features $\mathbf{f}_{i}$, and output HRMS $\hat{\textbf{I}}^\text{HR}_\text{MS}$ for improved learning.

\subsection{Limitations}

\noindent\textbf{Inference time limitations.}\quad Despite U-Know-DiffPAN's superior performance across multiple metrics, one notable limitation lies in its inference speed, compared to non-diffusion-based models. Diffusion models unavoidably require multiple sampling steps during inference, making this process inevitably slower. This iterative approach, while being beneficial for image generation of high quality, limits the model's efficiency in real-time applications. Future work could focus on optimizing the inference process by pruning sampling steps, while maintaining performance. 
\section{Conclusion}
We introduced U-Know-DiffPAN, a novel diffusion-based PAN-sharpening framework that effectively leverages frequency information from PAN and LRMS inputs to generate high-quality HRMS images. Our Uncertainty-aware Knowledge Distillation (U-Know) strategy enables efficient feature transfer by distilling both frequency-rich features and uncertainty maps from the teacher model, enhancing the restoration performance for spatially weak regions while reducing computational costs for the student model (FSA-S). Extensive experiments on multiple datasets, including WV3, QB, and GF2, demonstrate that our U-Know-DiffPAN outperforms very recent state-of-the-art methods in PAN-sharpening, establishing a new benchmark for performance and efficiency in satellite image processing.
\appendix

\section{More Details of Methods}
\subsection{Details of Vector Extractor}

\begin{figure}[h]
\centering{\includegraphics[width=\columnwidth]{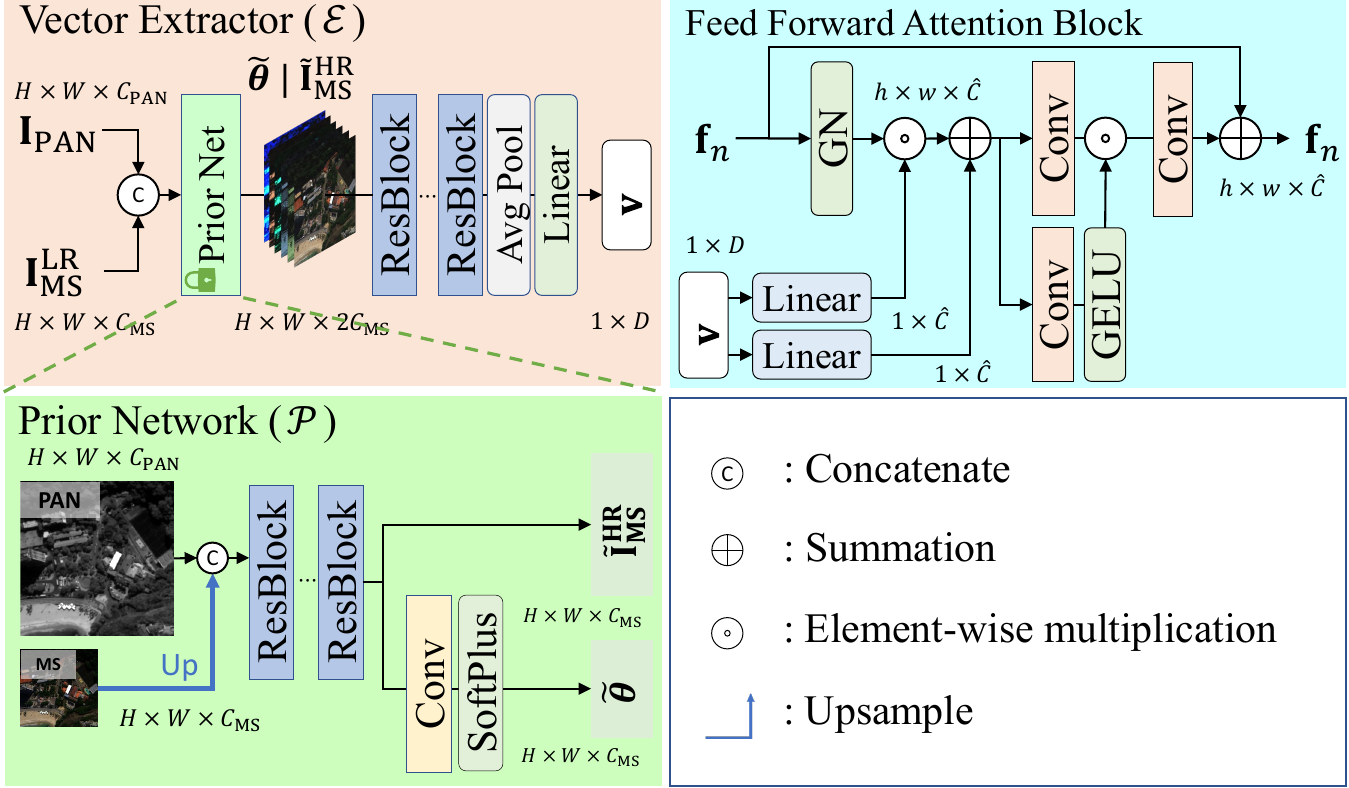}}
\caption[VE]{Detailed structure of Vector Extractor ($\mathcal{E}$). The simple representation of $\mathcal{E}$ is presented in Fig.~\ref{Method/FSA_T} of the main paper.}
\label{suppl/vectorextractor}
\end{figure}

The detailed structure of the Vector Extractor $\mathcal{E}$ in Fig. ~\ref{Method/FSA_T} of the main paper is depicted in Fig.~\ref{suppl/vectorextractor} and Eq. \ref{eq:suppl/vectorextractor}, which extracts an compact vector representation $\textbf{v} \in \mathbb{R}^{1 \times D}$ from two inputs: $\textbf{I}_\text{PAN}$ and $\textbf{I}^\text{LR}_\text{MS}$. $\mathcal{E}$ leverages a pretrained lightweight prior network $\mathcal{P}$ that takes $\textbf{I}_\text{PAN}$ and $\textbf{I}^\text{LR}_\text{MS}$ as inputs to generate a prior HRMS $\Tilde{\textbf{I}}^\text{HR}_\text{MS}$ and a prior uncertainty map $\boldsymbol{\Tilde{\theta}}$. These outputs, $\Tilde{\textbf{I}}^\text{HR}_\text{MS}$ and $\boldsymbol{\Tilde{\theta}}$, are then processed through multiple ResBlocks, followed by average pooling and a linear layer, to produce the final compact vector representation $\textbf{v}$. 
\begin{equation}
    \begin{aligned}
    &\left[\tilde{\boldsymbol{\theta}} \mid {\tilde{\textbf{I}}^\text{HR}_\text{MS}}\right] = \mathcal{P}\left(\textbf{I}_\text{PAN}, \textbf{I}^\text{LR}_\text{MS}\right), \\
    &\textbf{v} = \mathsf{Linear}\left(\mathsf{AvgPool}\left(\mathsf{ResBlock}^n\left(\left[\tilde{\boldsymbol{\theta}} \mid {\tilde{\textbf{I}}^\text{HR}_\text{MS}}\right]\right)\right)\right).
    \end{aligned}
    \label{eq:suppl/vectorextractor}
\end{equation}
where $\tilde{\boldsymbol{\theta}}$ denotes prior uncertainty map and $\tilde{\textbf{I}}^\text{HR}_\text{MS}$ is prior HRMS obtained from prior network $\mathcal{P}$. The compact vector representation $\textbf{v}$ not only encapsulates the combined information from $\textbf{I}_\text{PAN}$ and $\textbf{I}^\text{LR}_\text{MS}$, but also integrates the uncertainty information $\boldsymbol{\Tilde{\theta}}$ derived from the prior network $\mathcal{P}$. This representation is subsequently used as a conditioning input to the Feed Forward Attention (FFA) block in the encoder blocks of FSA-T.

\subsection{Fourier Transform Channel Attention (FTCA)}

\begin{figure}[h]
\centering{\includegraphics[width=\columnwidth]{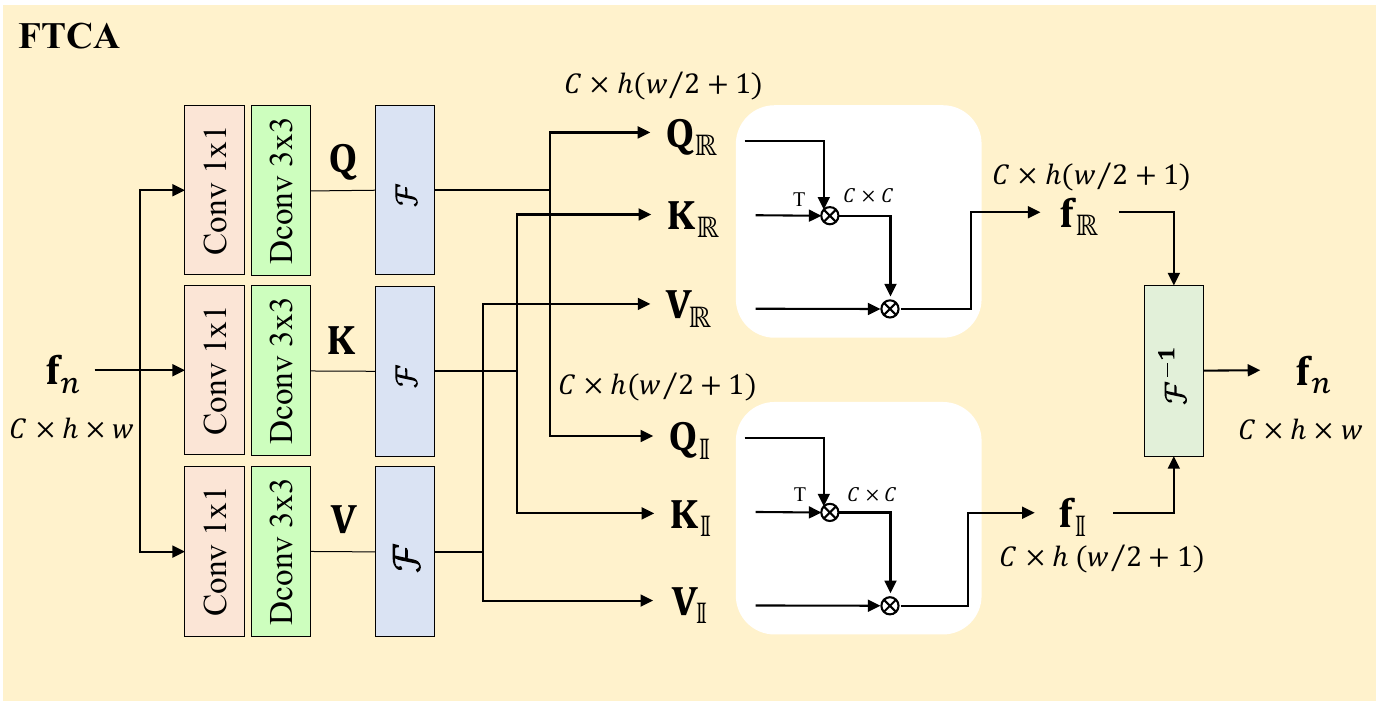}}
\caption[FTCA]{Detailed structure of Fourier Transform Channel Attention (FTCA) block. The simple representation of FTCA is presented in Fig.~\ref{Method/FSA_T} of the main paper.}
\label{suppl/FTCA}
\end{figure}

The detailed structure of FTCA in Fig. \ref{Method/FSA_T} of the main paper is depicted in Fig.~\ref{suppl/FTCA}, which is designed to enhance frequency domain features by applying the Discrete Fourier Transform (DFT) to feature maps, allowing self-attention to be performed more effectively in the frequency domain. As shown in Fig.~\ref{suppl/FTCA}, the input feature map is first transformed from the time domain to the frequency domain using 2D DFT. The real and imaginary parts of the 2D DFT feature map are then processed separately using channel attention. After applying channel attention to both real and imaginary parts, the Inverse Fourier Transform is applied to bring the features back into the time domain, producing the enhanced output. This transform-domain self attention facilitates more effective computations of frequency components to selectively emphasize. This process ensures that high-quality frequency information be effectively captured and utilized, resulting in enhanced feature representation and improved model performance.

\subsection{Stationary Wavelet Transform Cross Attention (SWTCA)}

\begin{figure}[h]
    \centering{\includegraphics[width=\columnwidth]{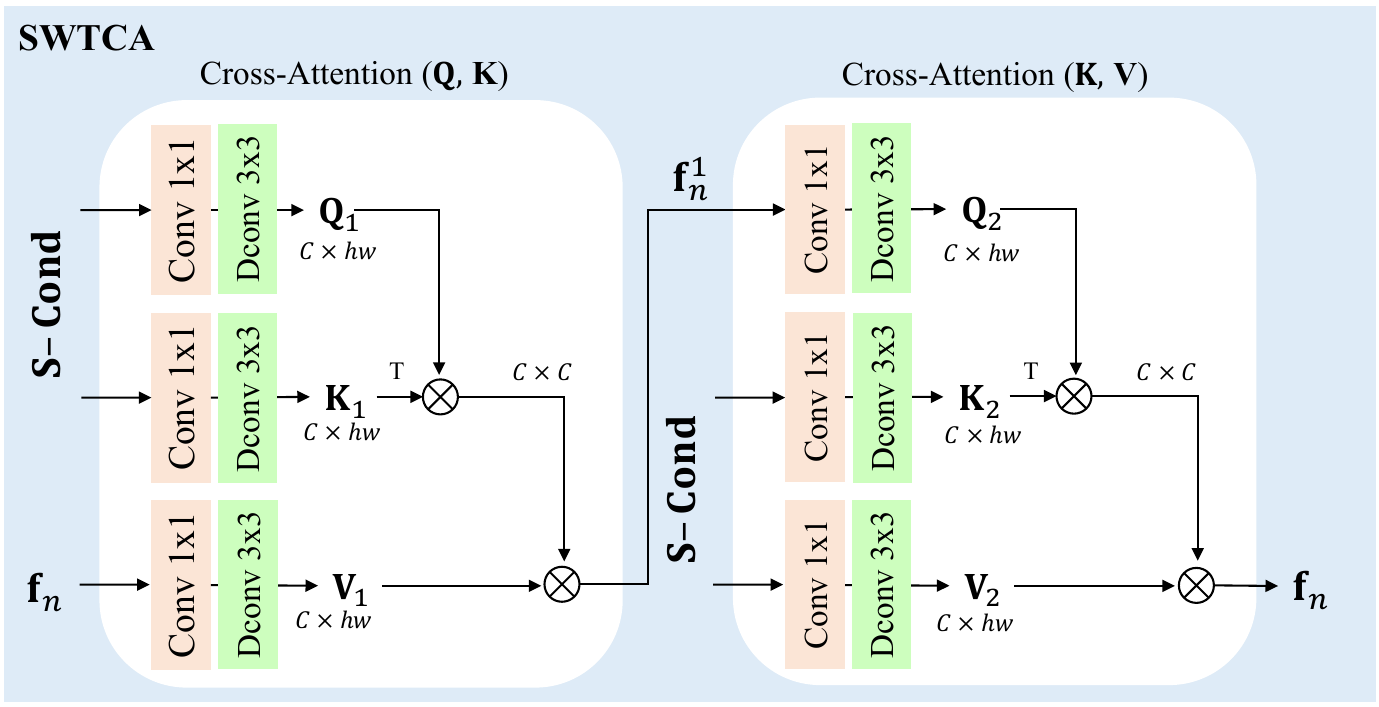}}
    \caption[SWTCA]{Detailed structure of our Stationary Wavelet Transform Cross Attention (SWTCA) block. The simple representation of SWTCA is presented in Fig.~\ref{Method/FSA_T} of the main paper.}
    \label{supple/SWTCA}
\end{figure}

\begin{figure*}[thp]
    \centering{\includegraphics[width=0.8\textwidth]{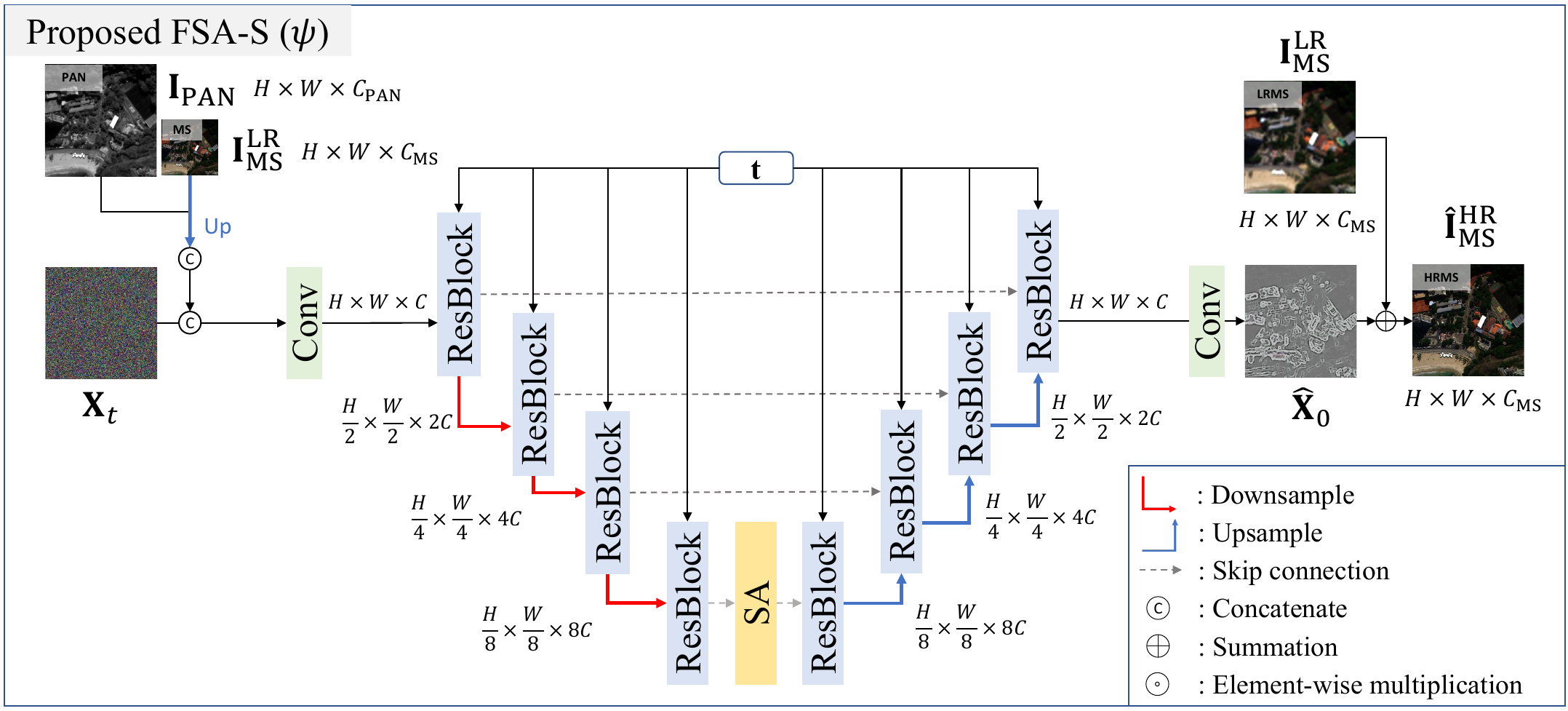}}
    \caption[SWTCA]{Detailed structure of FSA-S $\psi$.}
    \label{Supple/FSA-S}
\end{figure*}

Following the approach of ResDiff \cite{shang2024resdiff}, WaveDiff \cite{phung2023wavelet}, and WINet \cite{zhang2024pan}, the SWTCA in Fig. \ref{Method/FSA_T} of the main paper is detailed in Fig. \ref{supple/SWTCA}, which utilizes wavelet components of PAN and MS images decomposed using the Stationary Wavelet Transform (SWT). The SWTCA block is designed to inject additional frequency information into the features $\textbf{f}_n$ enhanced by the preceding FTCA. It selectively incorporates useful frequency components derived from the $\textbf{I}_\text{PAN}$ and $\textbf{I}^\text{LR}_\text{MS}$ using the SWT while maintaining shift invariance as conditions for our Diffusion Model $\Psi$ . The outputs of SWT for $\textbf{I}_\text{PAN}$ and $\textbf{I}^\text{LR}_\text{MS}$ are constructed respectively as follows:
\begin{equation}
    \begin{aligned}
    &\mathcal{S}(\mathbf{I}_\text{PAN}) = \left[\mathbf{L}_\text{PAN} \mid \mathbf{H}_\text{PAN} \mid \mathbf{V}_\text{PAN} \mid \mathbf{D}_\text{PAN}\right], \\
    &\mathcal{S}(\mathbf{I}_\text{MS}^\text{LR}) = \left[\mathbf{L}_\text{MS}^\text{LR} \mid \mathbf{H}_\text{MS}^\text{LR} \mid \mathbf{V}_\text{MS}^\text{LR} \mid \mathbf{D}_\text{MS}^\text{LR}\right],
    \end{aligned}
\end{equation}
where $\mathbf{L}$ represent the low-frequency approximation components, while $\mathbf{H}$, $\mathbf{V}$, $\mathbf{D}$ denote the high-frequency horizontal, vertical, and diagonal details, respectively. Since $\mathbf{I}_\text{PAN}$ provides highly detailed texture from its high spatial resolution but lacks spectral information, only its high-frequency components, which include $\mathbf{H}_\text{PAN}$, $\mathbf{V}_\text{PAN}$ and $\mathbf{D}_\text{PAN}$, are used. Conversely, $\mathbf{L}_\text{MS}^\text{LR}$ from $\mathbf{I}_\text{MS}^\text{LR}$ contains richer spectral information. So, these components are concatenated channel-wise to form $\mathcal{S}\text{-}\mathsf{Cond}$ that is the condition to be inputted to $\Psi$. $\mathcal{S}\text{-}\mathsf{Cond}$ is then constructed as follows:
\begin{equation}
    \mathcal{S}\text{-}\mathsf{Cond} = \left[\mathbf{L}_\text{MS}^\text{LR} \mid \mathbf{H}_\text{PAN} \mid \mathbf{V}_\text{PAN} \mid \mathbf{D}_\text{PAN}\right].
    \label{eq:S_Cond}
\end{equation}
Note that the construction of this kind condition is motivated from the traditional MRA(Multi-Resolution Analysis)-based \cite{otazu2005introduction, aiazzi2006mtf, zhou1998wavelet} PAN-sharpening methods that have utilized high-frequency Wavelet components of $\textbf{I}_\text{PAN}$ and low-frequency Wavelet components of $\textbf{I}^\text{LR}_\text{MS}$.
To inject $\mathcal{S}\text{-}\mathsf{Cond}$ into the feature $\mathbf{f}_{n}$ obtained from FTCA at the $n$-th decoder block, we employ a two-stage cross-attention process. In the first stage, $\mathbf{Q}_{1}$ and $\mathbf{K}_{1}$ are derived from $\mathcal{S}\text{-}\mathsf{Cond}$, directing attention toward important frequency components, while $\mathbf{V}_{1}$ is derived from $\mathbf{f}_{n}$. The resulting intermediate feature $\mathbf{f}_{n}^{1} \in \mathbb{R}^{C \times h \times w}$ is given as:
\begin{equation}
    \begin{aligned}
    &\left[\mathbf{Q}_{1} \mid \mathbf{K}_{1}\right] = \mathsf{Conv} (\mathcal{S}\text{-}\mathsf{Cond}),\;\mathbf{V}_{1} = \mathsf{Conv}(\mathbf{f}_{n}), \\
    &\mathbf{f}_{n}^{1} = \mathsf{SoftMax}\left(\mathbf{Q}_{1} \mathbf{K}_{1}^{\mathsf{T}}/{\sqrt{C}}\right) \mathbf{V}_{1},
    \end{aligned}
\end{equation}
where $\mathbf{Q}_{1}$, $\mathbf{K}_{1}$, $\mathbf{V}_{1} \in \mathbb{R}^{C \times hw}$ are a Query, a Key and a Value, respectively. The $\mathbf{Q}_{1}$, $\mathbf{K}_{1}$ and $\mathbf{V}_{1}$ are reshaped to \( C \times hw \) for channel attention. In the second stage, $\mathbf{Q}_{2}$ is derived from the intermediate feature $\mathbf{f}_{n}^{1}$, while $\mathbf{K}_{2}$ and $\mathbf{V}_{2}$ come from $\mathcal{S}\text{-}\mathsf{Cond}$, enabling the model to refine the correlation between the reconstructed feature $\mathbf{f}_{n}$ and the frequency components in $\mathcal{S}\text{-}\mathsf{Cond}$. This stage yields the final feature $\mathbf{f}_{n} \in \mathbb{R}^{C \times h \times w}$ after reshaping $\mathbf{Q}_{2}$, $\mathbf{K}_{2}$ and $\mathbf{V}_{2}$ into \( C \times hw \) for channel attention, which is given as follows:
\begin{equation}
    \begin{aligned}
    &\mathbf{Q}_{2} = \mathsf{Conv}(\mathbf{f}_{n}^{1}), \; \left[\mathbf{K}_{2} \mid \mathbf{V}_{2}\right] = \mathsf{Conv}(\mathcal{S}\text{-}\mathsf{Cond}), \\
    &\mathbf{f}_{n} \leftarrow \mathsf{SoftMax}\left(\mathbf{Q}_{2} \mathbf{K}_{2}^{\mathsf{T}}/{\sqrt{C}}\right) \mathbf{V}_{2}.
    \end{aligned}
\end{equation}
The first cross-attention block in Fig. \ref{supple/SWTCA} uses both Query and Key mappings from $\mathcal{S}\text{-}\mathsf{Cond}$, focusing the attention map purely on $\mathcal{S}\text{-}\mathsf{Cond}$ information, and aiding in concentrating on the significant channels of the feature $ \textbf{f}_n $. The second cross-attention block in Fig. \ref{supple/SWTCA} uses Query from the intermediate feature $ \textbf{f}^{1}_n $ and Key from $\mathcal{S}\text{-}\mathsf{Cond}$, learning the correlation between the reconstructed feature $ \textbf{f}^{1}_n $ and the meaningful frequency information in $\mathbf{I}_\text{PAN}$ and $\mathbf{I}^\text{LR}_\text{MS}$. This block helps emphasizing significant frequency components within $\mathcal{S}\text{-}\mathsf{Cond}$.

\subsection{Student network FSA-S ($\psi$) architecture}

The student model, denoted as FSA-S $\psi$, is a diffusion-based model designed with the same number of encoder and decoder blocks as FSA-T $\Psi$. Each block consists solely of ResBlocks, and unlike the teacher $\Psi$, the FSA-S operates without any additional conditional inputs. As shown in Fig.\ref{Method/kd_model}, the lightweight FSA-S $\psi$ predicts denoised image $\widetilde{\mathbf{X}}_{0}$ from input $\mathbf{X}_{t}$ at timestep $t$ as:
\begin{equation}
    \widetilde{\mathbf{X}}_{0} = \psi\left(\left[\mathbf{X}_{t}\mid\mathbf{I}_\text{PAN}\mid\mathbf{I}_\text{MS}^\text{LR}\right]; t\right).
\end{equation}

\section{Stationary Wavelet Transform compared to Discrete Wavelet Transform}
Discrete Wavelet transforms have been widely used in image processing and PAN-sharpening tasks due to their ability to analyze signals across multiple resolutions. Discrete Wavelet Transform (DWT) decomposes signals into low-frequency (approximation) and high-frequency (detail) components at each level. Although DWT has proven to be effective in numerous applications, it suffers from shift variance due to its inherent downsampling operation at each decomposition level. To address this limitation, the Stationary Wavelet Transform (SWT) was introduced. Unlike DWT, SWT omits the downsampling step, ensuring that the signal size remains constant across all levels. This design enables SWT to maintain shift invariance, meaning that minor shifts in the input signal do not affect the transformation results. This property makes SWT highly advantageous in applications such as PAN-sharpening, where maintaining consistency in transformed results is crucial. Specifically, SWT ensures that the wavelet coefficients remain stable even when the input signal undergoes small translations, leading to more robust performance in PAN-sharpening tasks. In addition, SWT allows for higher resolution analysis across all levels, as the image or signal is not downsampled during the transformation. Compared to DWT that inherently reduces the image resolutions at each level, SWT retains the original resolution, making it more suitable for tasks that require detailed frequency information. 

\subsection{SWT condition versus DWT condition}

\begin{table}[htbp]
    \scriptsize
    \centering
    \resizebox{0.99\columnwidth}{!}{
    \def\arraystretch{1.2}
    \begin{tabular} {c|c|c|c|c}
    \Xhline{2\arrayrulewidth}
     & \multicolumn{4}{c}{\textbf{GF2} Dataset (Reduced-Resolution)} \\
    \hline
    Condition & SAM$\downarrow$ & ERGAS$\downarrow$ & SCC$\uparrow$ & Q4$\uparrow$\\
    \hline
    DWT & 0.646 $\pm$ 0.117 & 0.567 $\pm$ 0.095 & 0.993 $\pm$ 0.002 & 0.987 $\pm$ 0.007 \\
    SWT & {\color{red}{\textbf{0.603 $\pm$ 0.102}}} & {\color{red}{\textbf{0.537 $\pm$ 0.077}}} & {\color{red}{\textbf{0.994 $\pm$ 0.001}}} & {\color{red}{\textbf{0.988 $\pm$ 0.006}}} \\
    \Xhline{2\arrayrulewidth}
    \end{tabular}}
    \caption{Comparison of results between DWT and SWT conditioning at the SWTCA block in FSA-T $\Psi$, with the best values highlighted in red.}
  \label{tab:SWT vs DWT}
\end{table}

Table.~\ref{tab:SWT vs DWT} represents the performance differences between DWT and SWT for the $\mathcal{S}\text{-}\mathsf{Cond}$ in Eq.~\ref{eq:S_Cond} as the conditioning input to the SWTCA block in FSA-T $\Psi$. The SWT-based $\mathcal{S}\text{-}\mathsf{Cond}$ outperforms the DWT-based one across all metrics. The results in Table.~\ref{tab:SWT vs DWT} highlight that SWT achieves lower SAM and ERGAS values, as well as higher SCC and Q4 scores, indicating its superior performance in PAN-sharpening tasks. This improvement underscores the advantages of shift invariance and resolution preservation provided by SWT.

\section{Uncertainty Estimation}
Uncertainty estimation has been an important topic of research in deep learning. Several works \cite{van2020uncertainty, lakshminarayanan2017simple, ashukha2020pitfalls, seitzer2022pitfalls} have incorporated uncertainty into regression problems. A Bayesian deep learning framework was proposed to enable its application to per-pixel computer vision tasks. Similarly, some other works \cite{seitzer2022pitfalls, kendall2017uncertainties, chang2020data} explored the role of data uncertainty by modeling both the mean and variance of the predictions. Uncertainty-based loss function of those works can be represented as:
\begin{equation}
\mathcal{L} = \frac{1}{N} \sum_{i=1}^N \frac{\|\textbf{x}_i - f(\textbf{y}_i)\|_2}{2 \boldsymbol{\sigma}_i^2} + \frac{1}{2} \ln \boldsymbol{\sigma}_i^2
\end{equation}
where $N$ denotes total number of input samples, $\textbf{x}_i$ is a target label, $\textbf{y}_i$ is an input, $f(\textbf{y}_i)$ and $\boldsymbol{\sigma}^2_i$ denote the learned mean and variance, respectively. Recent studies \cite{ning2021uncertainty, seitzer2022pitfalls} have continued to explore uncertainty estimation in deep learning, particularly in a task requiring high accuracy such as image super-resolution \cite{ning2021uncertainty}. These approaches have demonstrated that such uncertainty-based losses that utilize uncertainty terms can achieve better results than mean square error (MSE) and mean absolute error (MAE) losses. Inspired by this work, we formulated our $\mathcal{L}_\text{U-Diff}$ as:
\begin{equation}
    \mathcal{L}_\text{U-Diff} = \left\lVert \frac{1}{2\hat{\bm{\theta}}} \odot \left| \widehat{\mathbf{X}}_0 - \mathbf{X}_0 \right|+ \frac{1}{2}\log \hat{\bm{\theta}} \right\rVert_1,
    \label{eq:U-diff_suppl}
\end{equation}
where $\widehat{\mathbf{X}}_0$ is a predicted residual, ${\mathbf{X}}_0$ is a target, and $\hat{\bm{\theta}}$ serves as the estimated variance term and is regarded as the uncertainty map in our framework. In PAN-sharpening tasks, the regions with high uncertainty, such as complex textures and edges, are visually more significant than smooth areas. By prioritizing these regions, uncertainty-aware models can handle complex image details more effectively, potentially leading to improved PAN-sharpening performance in our U-Know-DiffPAN framework.

\section{Additional Results}

\begin{table*}[tbp]
\centering
\resizebox{0.99\textwidth}{!}{ 
\begin{tabular}{|c|cccccc|ccc|}
\hline
 \textbf{WV3}& \multicolumn{6}{c}{Reduced-Resolution}& \multicolumn{3}{c|}{Full-Resolution} \\
\hline
\textbf{Model}  & \textbf{PSNR$\uparrow$}  & \textbf{SSIM$\uparrow$}   & \textbf{SAM$\downarrow$}    & \textbf{ERGAS$\downarrow$}  & \textbf{SCC$\uparrow$} & \textbf{Q8$\uparrow$}& $\textbf{D}_\lambda$$\downarrow$ & $\textbf{D}_s$$\downarrow$ & \textbf{HQNR}$\uparrow$ \\
\hline
PanNet\cite{yang2017pannet}& 36.148 $\pm$ 1.958& 0.966 $\pm$ 0.011
& 3.402 $\pm$ 0.672 & 2.538 $\pm$ 0.597 & 0.979 $\pm$ 0.006 & 0.913 $\pm$ 0.087& 0.035 $\pm$ 0.014& 0.049 $\pm$ 0.019& 0.918 $\pm$ 0.031\\
MSDCNN\cite{yuan2018multiscale}& 36.329 $\pm$ 1.748& 0.967 $\pm$ 0.010
& 3.300 $\pm$ 0.654 & 2.489 $\pm$ 0.620 & 0.979 $\pm$ 0.007 & 0.914 $\pm$ 0.087& 0.028 $\pm$ 0.013& 0.050 $\pm$ 0.020& 0.924 $\pm$ 0.030\\
FusionNet\cite{wu2021dynamic}& 36.569 $\pm$ 1.666& 0.968 $\pm$ 0.009
& 3.188 $\pm$ 0.628 & 2.428 $\pm$ 0.621 & 0.981 $\pm$ 0.007 & 0.916 $\pm$ 0.087& 0.029 $\pm$ 0.011& 0.053 $\pm$ 0.021& 0.920 $\pm$ 0.030\\
LAGNet\cite{jin2022lagconv}& 36.732 $\pm$ 1.723& 0.970 $\pm$ 0.009
& 3.153 $\pm$ 0.608 & 2.380 $\pm$ 0.617 & 0.981 $\pm$ 0.007 & 0.916 $\pm$ 0.087& 0.033 $\pm$ 0.012& 0.055 $\pm$ 0.023& 0.915 $\pm$ 0.033\\
S2DBPN\cite{zhang2023spatial}& 37.216 $\pm$ 1.888& 0.972 $\pm$ 0.009
& 3.019 $\pm$ 0.588 & 2.245 $\pm$ 0.541 & 0.985 $\pm$ 0.005 & 0.917 $\pm$ 0.091& 0.025 $\pm$ 0.010& 0.030 $\pm$ 0.010 & 0.946 $\pm$ 0.018\\
DCPNet\cite{zhang2024dcpnet}& 37.009 $\pm$ 1.735& 0.972 $\pm$ 0.008
& 3.083 $\pm$ 0.537 & 2.301 $\pm$ 0.569 & 0.984 $\pm$ 0.005 & 0.915 $\pm$ 0.092& 0.043 $\pm$ 0.018& 0.036 $\pm$ 0.012& 0.923 $\pm$ 0.027\\
CANConv\cite{duan2024content}   &  37.441 $\pm$ 1.788&  0.973 $\pm$ 0.008& 2.927 $\pm$ 0.536& 2.163 $\pm$ 0.481& 0.985 $\pm$ 0.005& 0.918 $\pm$ 0.082& 0.020 $\pm$ 0.008& {\color{blue}{\underline{0.030 $\pm$ 0.008}}}& 0.951 $\pm$ 0.013
\\
PanDiff\cite{meng2023pandiff}& 37.029 $\pm$ 1.796& 0.971 $\pm$ 0.008&  3.058 $\pm$ 0.567 &  2.276 $\pm$ 0.545 &  0.984 $\pm$ 0.004 &  0.913 $\pm$ 0.084& {\color{red}{\textbf{0.014 $\pm$ 0.005}}}& 0.034 $\pm$ 0.005& 0.952 $\pm$ 0.009
\\
TMDiff\cite{xing2024empower}&  37.477 $\pm$ 1.923&  {\color{blue}{\underline{0.973 $\pm$ 0.008}}}&  2.885 $\pm$ 0.549&  2.151 $\pm$ 0.458&  0.986 $\pm$ 0.004&  0.915 $\pm$ 0.086& 0.018 $\pm$ 0.007& 0.059 $\pm$ 0.009& 0.924 $\pm$ 0.015\\
\cellcolor{blue!15}\textbf{FSA-T}   & \cellcolor{blue!15}{\color{blue}{\underline{37.894 $\pm$ 1.820}}}& \cellcolor{blue!15}{\color{red}{\textbf{0.976 $\pm$ 0.007}}}& \cellcolor{blue!15}{\color{blue}{\underline{2.801 $\pm$ 0.517}}}& \cellcolor{blue!15}{\color{blue}{\underline{2.055 $\pm$ 0.463}}}& \cellcolor{blue!15}{\color{blue}{\underline{0.987 $\pm$ 0.003}}}& \cellcolor{blue!15}{\color{blue}{\underline{0.921 $\pm$ 0.083}}}& \cellcolor{blue!15}{\color{red}{\textbf{0.014 $\pm$ 0.005}}}& \cellcolor{blue!15}0.032 $\pm$ 0.003& \cellcolor{blue!15}{\color{blue}{\underline{0.954 $\pm$ 0.006}}}
\\
\cellcolor{blue!15}\textbf{FSA-S}   & \cellcolor{blue!15}{\color{red}{\textbf{37.930 $\pm$ 1.824}}}& \cellcolor{blue!15}{\color{red}{\textbf{0.976 $\pm$ 0.007}}}& \cellcolor{blue!15}{\color{red}{\textbf{2.797 $\pm$ 0.526}}}& \cellcolor{blue!15}{\color{red}{\textbf{2.046 $\pm$ 0.454}}}& \cellcolor{blue!15}{\color{red}{\textbf{0.988 $\pm$ 0.003}}}& \cellcolor{blue!15}{\color{red}{\textbf{0.922 $\pm$ 0.083}}}& \cellcolor{blue!15}{\color{blue}{\underline{0.016 $\pm$ 0.006}}}& \cellcolor{blue!15}{\color{red}{\textbf{0.029 $\pm$ 0.003}}}& \cellcolor{blue!15}{\color{red}{\textbf{0.955 $\pm$ 0.008}}}
\\
\hline

\textbf{QB}& \multicolumn{6}{c}{Reduced-Resolution}& \multicolumn{3}{c|}{Full-Resolution} \\
\hline
\textbf{Model}  & \textbf{PSNR$\uparrow$}  & \textbf{SSIM$\uparrow$}   & \textbf{SAM$\downarrow$}    & \textbf{ERGAS$\downarrow$}  & \textbf{SCC$\uparrow$} & \textbf{Q4$\uparrow$}   & $\textbf{D}_\lambda$$\downarrow$& $\textbf{D}_s$$\downarrow$& \textbf{HQNR}$\uparrow$\\
\hline
PanNet\cite{yang2017pannet}& 35.563 $\pm$ 1.930& 0.939 $\pm$ 0.012
& 5.273 $\pm$ 0.946& 4.856 $\pm$ 0.590& 0.966 $\pm$ 0.015& 0.911 $\pm$ 0.094& 0.063 $\pm$ 0.019& 0.092 $\pm$ 0.021& 0.851 $\pm$ 0.035
\\
MSDCNN\cite{yuan2018multiscale}& 37.040 $\pm$ 1.778& 0.954 $\pm$ 0.007
& 4.828 $\pm$ 0.824& 4.074 $\pm$ 0.244& 0.977 $\pm$ 0.010& 0.925 $\pm$ 0.098& 0.058 $\pm$ 0.014& 0.058 $\pm$ 0.027& 0.888 $\pm$ 0.037
\\
FusionNet\cite{wu2021dynamic}& 36.821 $\pm$ 1.765& 0.952 $\pm$ 0.007
& 4.892 $\pm$ 0.822& 4.183 $\pm$ 0.266& 0.975 $\pm$ 0.011& 0.923 $\pm$ 0.100& 0.074 $\pm$ 0.022& 0.079 $\pm$ 0.025& 0.853 $\pm$ 0.041
\\
LAGNet\cite{jin2022lagconv}& 37.565 $\pm$ 1.721& 0.958 $\pm$ 0.006
& 4.682 $\pm$ 0.785& 3.845 $\pm$ 0.323& 0.980 $\pm$ 0.009& 0.930 $\pm$ 0.095& 0.075 $\pm$ 0.019& 0.035 $\pm$ 0.009& 0.892 $\pm$ 0.024
\\
S2DBPN\cite{zhang2023spatial}& 37.314 $\pm$ 1.782& 0.956 $\pm$ 0.006
& 4.849 $\pm$ 0.822& 3.956 $\pm$ 0.291& 0.980 $\pm$ 0.008& 0.928 $\pm$ 0.093& 0.059 $\pm$ 0.026& 0.036 $\pm$ 0.023& 0.908 $\pm$ 0.044
\\
DCPNet\cite{zhang2024dcpnet}& 38.079 $\pm$ 1.454& \color{blue}{\underline{0.963 $\pm$ 0.004}}
& 4.420 $\pm$ 0.710& 3.618 $\pm$ 0.313& 0.983 $\pm$ 0.010& 0.935 $\pm$ 0.095& 0.051 $\pm$ 0.017& 0.073 $\pm$ 0.013& 0.880 $\pm$ 0.013
\\
CANConv\cite{duan2024content}   &  37.795 $\pm$ 1.801
&  0.960 $\pm$ 0.006& 4.554 $\pm$ 0.788
& 3.740 $\pm$ 0.304& 0.982 $\pm$ 0.007& 0.935 $\pm$ 0.087& 0.039 $\pm$ 0.012& 0.070 $\pm$ 0.017& 0.893 $\pm$ 0.010
\\
PanDiff\cite{meng2023pandiff}& 37.842 $\pm$ 1.721
& 0.959 $\pm$ 0.006& 4.611 $\pm$ 0.768
& 3.723 $\pm$ 0.280& 0.982 $\pm$ 0.007& 0.935 $\pm$ 0.084& \color{red}{\textbf{0.028 $\pm$ 0.011}}& 0.055 $\pm$ 0.012& 0.919 $\pm$ 0.010
\\
TMDiff\cite{xing2024empower}&  37.642 $\pm$ 1.831&  0.958 $\pm$ 0.006& 4.627 $\pm$ 0.814& 3.804 $\pm$ 0.279& 0.981 $\pm$ 0.008& 0.930 $\pm$ 0.096& \color{blue}{\underline{0.034 $\pm$ 0.016}}& 0.068 $\pm$ 0.012& 0.901 $\pm$ 0.011\\
\cellcolor{blue!15}\textbf{FSA-T}   & \cellcolor{blue!15}\color{blue}{\underline{38.343 $\pm$ 1.718}}& \cellcolor{blue!15}\color{red}{\textbf{0.964 $\pm$ 0.005}}& \cellcolor{blue!15}\color{blue}{\underline{4.349 $\pm$ 0.723}}& \cellcolor{blue!15}\color{blue}{\underline{3.502 $\pm$ 0.272}}& \cellcolor{blue!15}\color{red}{\textbf{0.985 $\pm$ 0.007}}& \cellcolor{blue!15}\color{red}{\textbf{0.938 $\pm$ 0.089}}& \cellcolor{blue!15}0.036 $\pm$ 0.018& \cellcolor{blue!15}\color{red}{\textbf{0.031 $\pm$ 0.014}}& \cellcolor{blue!15}\color{red}{\textbf{0.934 $\pm$ 0.029}}
\\
\cellcolor{blue!15}\textbf{FSA-S}   & \cellcolor{blue!15}\color{red}{\textbf{38.361 $\pm$ 1.709}}
& \cellcolor{blue!15}\color{red}{\textbf{0.964 $\pm$ 0.005}}& \cellcolor{blue!15}\color{red}{\textbf{4.337 $\pm$ 0.733}}& \cellcolor{blue!15}\color{red}{\textbf{3.500 $\pm$ 0.272}}& \cellcolor{blue!15}\color{blue}{\underline{0.984 $\pm$ 0.007}}& \cellcolor{blue!15}\color{blue}{\underline{0.938 $\pm$ 0.090}}& \cellcolor{blue!15}0.035 $\pm$ 0.011& \cellcolor{blue!15}{\color{blue}{\underline{0.035 $\pm$ 0.021}}}& \cellcolor{blue!15}\color{blue}{\underline{0.931 $\pm$ 0.029}}
\\
\hline

\textbf{GF2}& \multicolumn{6}{c}{Reduced-Resolution}&\multicolumn{3}{c|}{Full-Resolution}\\
\hline
\textbf{Model}           & \textbf{PSNR$\uparrow$}  & \textbf{SSIM$\uparrow$}   & \textbf{SAM$\downarrow$}    & \textbf{ERGAS$\downarrow$}  & \textbf{SCC$\uparrow$}    & \textbf{Q4$\uparrow$}   & $\textbf{D}_\lambda$$\downarrow$& $\textbf{D}_s$$\downarrow$&\textbf{HQNR}$\uparrow$\\
\hline
PanNet\cite{yang2017pannet}      & 39.197 $\pm$ 2.009& 0.959 $\pm$ 0.011& 1.050 $\pm$ 0.209& 1.038 $\pm$ 0.214& 0.975 $\pm$ 0.006& 0.963 $\pm$ 0.009& 0.020 $\pm$ 0.012& 0.052 $\pm$ 0.009& 0.929 $\pm$ 0.013
\\
MSDCNN\cite{yuan2018multiscale}      & 40.730 $\pm$ 1.564& 0.971 $\pm$ 0.006& 0.946 $\pm$ 0.166& 0.862 $\pm$ 0.141& 0.983 $\pm$ 0.003& 0.972 $\pm$ 0.009& 0.026 $\pm$ 0.014& 0.079 $\pm$ 0.011& 0.898 $\pm$ 0.016
\\
FusionNet\cite{wu2021dynamic}   & 39.866 $\pm$ 1.955& 0.966 $\pm$ 0.009& 0.971 $\pm$ 0.195& 0.960 $\pm$ 0.193& 0.980 $\pm$ 0.005& 0.967 $\pm$ 0.008& 0.034 $\pm$ 0.013& 0.105 $\pm$ 0.013& 0.865 $\pm$ 0.018
\\
LAGNet\cite{jin2022lagconv}     & 41.147 $\pm$ 1.384& 0.974 $\pm$ 0.005& 0.886 $\pm$ 0.140& 0.816 $\pm$ 0.121& 0.985 $\pm$ 0.003& 0.974 $\pm$ 0.009& 0.030 $\pm$ 0.014& 0.078 $\pm$ 0.013& 0.895 $\pm$ 0.021
\\
S2DBPN\cite{zhang2023spatial}      & 42.686 $\pm$ 1.676& 0.980 $\pm$ 0.005& 0.772 $\pm$ 0.149& 0.686 $\pm$ 0.125& 0.990 $\pm$ 0.002& 0.981 $\pm$ 0.007& 0.020 $\pm$ 0.012& 0.046 $\pm$ 0.007& 0.935 $\pm$ 0.011
\\
DCPNet\cite{zhang2024dcpnet}      & 42.312 $\pm$ 1.682& 0.979 $\pm$ 0.005& 0.806 $\pm$ 0.153& 0.724 $\pm$ 0.138& 0.988 $\pm$ 0.003& 0.980 $\pm$ 0.007& 0.024 $\pm$ 0.022& \color{red}{\textbf{0.024 $\pm$ 0.008}}& {\color{red}{\textbf{0.953 $\pm$ 0.019}}}
\\
CANConv\cite{duan2024content}&  43.166 $\pm$ 1.705&  0.982 $\pm$ 0.004& 0.722 $\pm$ 0.138& 0.653 $\pm$ 0.124&  0.991 $\pm$ 0.002& 0.983 $\pm$ 0.006& 0.019 $\pm$ 0.010& 0.063 $\pm$ 0.009& 0.919 $\pm$ 0.011
\\
PanDiff\cite{meng2023pandiff}& 42.827 $\pm$ 1.462& 0.980 $\pm$ 0.005& 0.767 $\pm$ 0.134& 0.674 $\pm$ 0.110& 0.990 $\pm$ 0.002&  0.981 $\pm$ 0.007& 0.020 $\pm$ 0.014& 0.045 $\pm$ 0.009& 0.936 $\pm$ 0.011
\\
TMDiff\cite{xing2024empower}&  41.896 $\pm$ 1.765&  0.977 $\pm$ 0.005& 0.764 $\pm$ 0.155& 0.754 $\pm$ 0.143& 0.988 $\pm$ 0.003& 0.979 $\pm$ 0.007& 0.029 $\pm$ 0.011& {\color{blue}{\underline{0.030 $\pm$ 0.010}}}& 0.942 $\pm$ 0.016\\
\cellcolor{blue!15}\textbf{FSA-T}          & \cellcolor{blue!15}{\color{red}{\textbf{44.757 $\pm$ 1.359}}}& \cellcolor{blue!15}{\color{red}{\textbf{0.988 $\pm$ 0.003}}}& \cellcolor{blue!15}{\color{red}{\textbf{0.603 $\pm$ 0.102}}}& \cellcolor{blue!15}{\color{red}{\textbf{0.537 $\pm$ 0.077}}}& \cellcolor{blue!15}{\color{red}{\textbf{0.994 $\pm$ 0.001}}}& \cellcolor{blue!15}{\color{red}{\textbf{0.988 $\pm$ 0.006}}}& \cellcolor{blue!15}{\color{red}{\textbf{0.017 $\pm$ 0.010}}}& \cellcolor{blue!15}{\color{blue}{\underline{0.030 $\pm$ 0.008}}}& \cellcolor{blue!15}{\color{red}{\textbf{0.953 $\pm$ 0.013}}}
\\
\cellcolor{blue!15}\textbf{FSA-S}          & \cellcolor{blue!15}{\color{blue}{\underline{44.585 $\pm$ 1.521}}}& \cellcolor{blue!15}{\color{blue}{\underline{0.986 $\pm$ 0.003}}}& \cellcolor{blue!15}{\color{blue}{\underline{0.624 $\pm$ 0.109}}}& \cellcolor{blue!15}{\color{blue}{\underline{0.548 $\pm$ 0.091}}}& \cellcolor{blue!15}{\color{blue}{\underline{0.993 $\pm$ 0.001}}}& \cellcolor{blue!15}{\color{blue}{\underline{0.987 $\pm$ 0.007}}}& \cellcolor{blue!15}{\color{blue}{\underline{0.018 $\pm$ 0.011}}}& \cellcolor{blue!15}0.037 $\pm$ 0.007& \cellcolor{blue!15}\color{blue}{\underline{0.944 $\pm$ 0.012}}
\\
\hline

\end{tabular}}
\caption{Additional PAN-sharpening results by our U-Know-DiffPAN and other SOTA methods for the WV3, QB, and GF2 dataset. The best (second best) performance in each block is highlighted in bold {\color{red}{\textbf{red}}} (underlined in {\color{blue}{\underline{blue}}}). }
\label{table:WV3,QB,GF2 total result}
\end{table*}

\begin{figure*}[tbp]
    \centering{\includegraphics[width=0.99\textwidth]{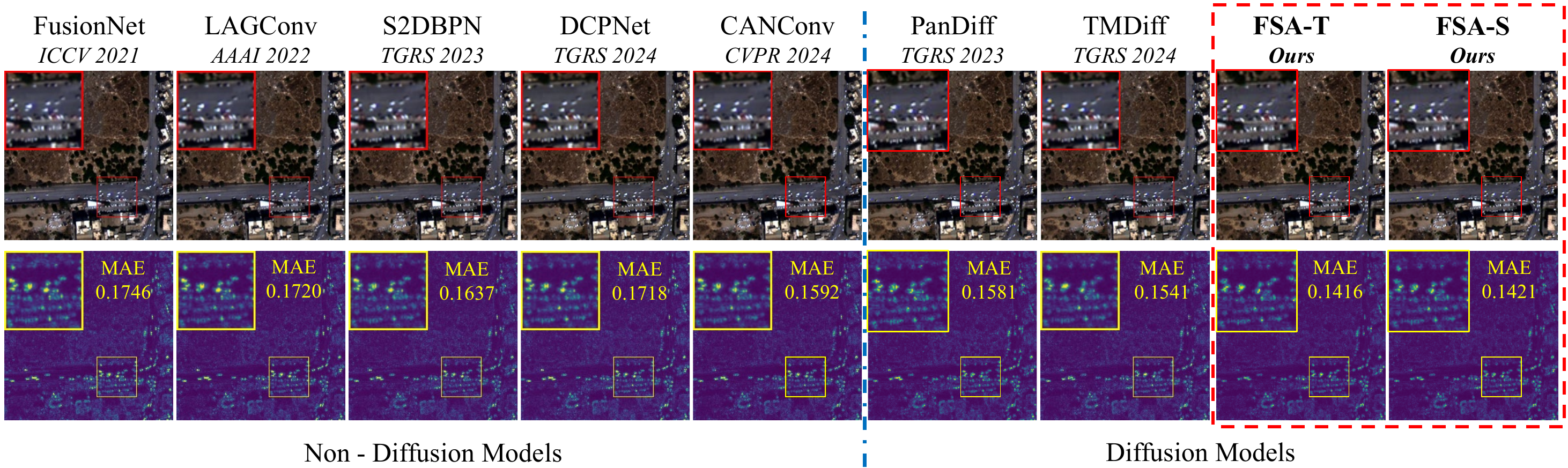}}
    \caption[RR_WV3]{PAN-sharpening results for the WV3 dataset under reduced resolution (RR) scenarios. The first row depicts the output HRMS images, while the second row highlights the Error Map between the output HRMS and the corresponding ground truth images.The Mean Absolute Error (MAE) values are presented alongside the Error Map. Zoom in for better visualization.}
    \label{Supple/RR_WV3}
\end{figure*}

\begin{figure*}[tbp]
    \centering{\includegraphics[width=0.99\textwidth]{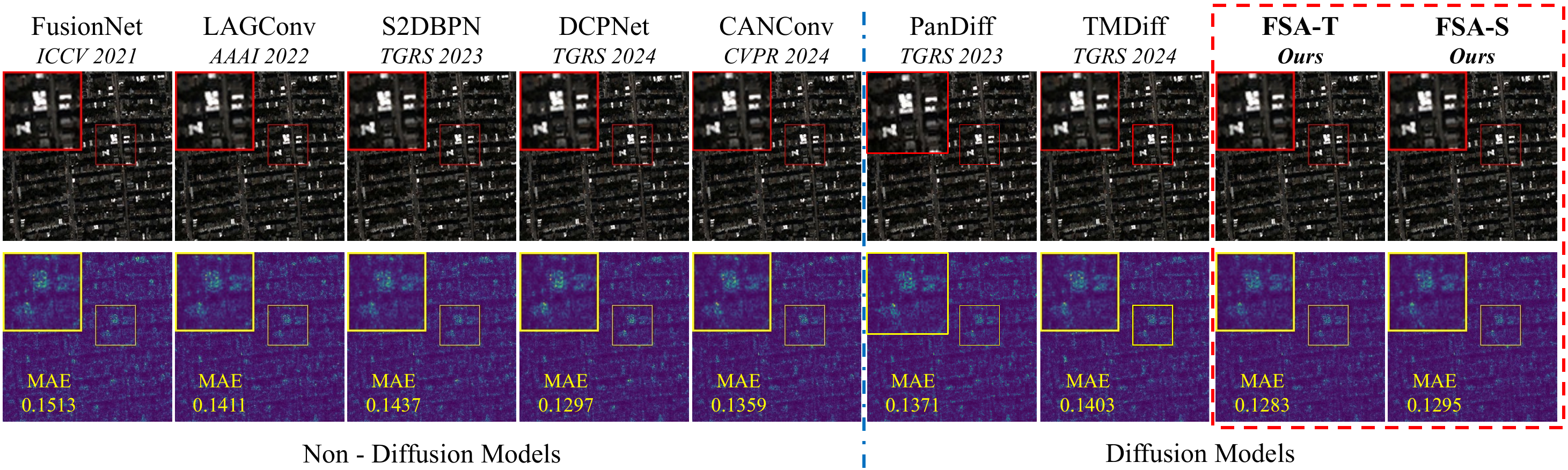}}
    \caption[RR_QB]{PAN-sharpening results for the QB dataset under reduced resolution (RR) scenarios. The first row depicts the output HRMS images, while the second row highlights the Error Map between the output HRMS and the corresponding ground truth images. The Mean Absolute Error (MAE) values are presented alongside the Error Map. Zoom in for better visualization.}
    \label{Supple/RR_QB}
\end{figure*}

\begin{figure*}[tbp]
    \centering{\includegraphics[width=0.99\textwidth]{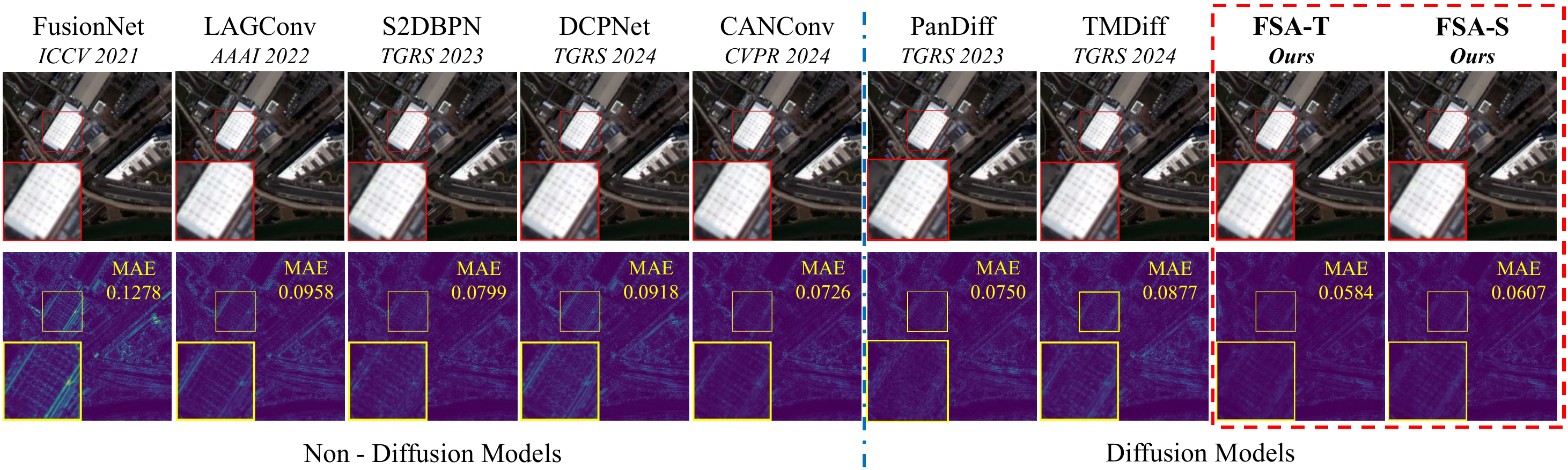}}
    \caption[RR_GF2]{PAN-sharpening results for the GF2 dataset under reduced resolution (RR) scenarios. The first row depicts the output HRMS images, while the second row highlights the Error Map between the output HRMS and the corresponding ground truth images. The Mean Absolute Error (MAE) values are presented alongside the Error Map. Zoom in for better visualization.}
    \label{Supple/RR_GF2}
\end{figure*}

\begin{figure*}[tbp]
    \centering{\includegraphics[width=0.99\textwidth]{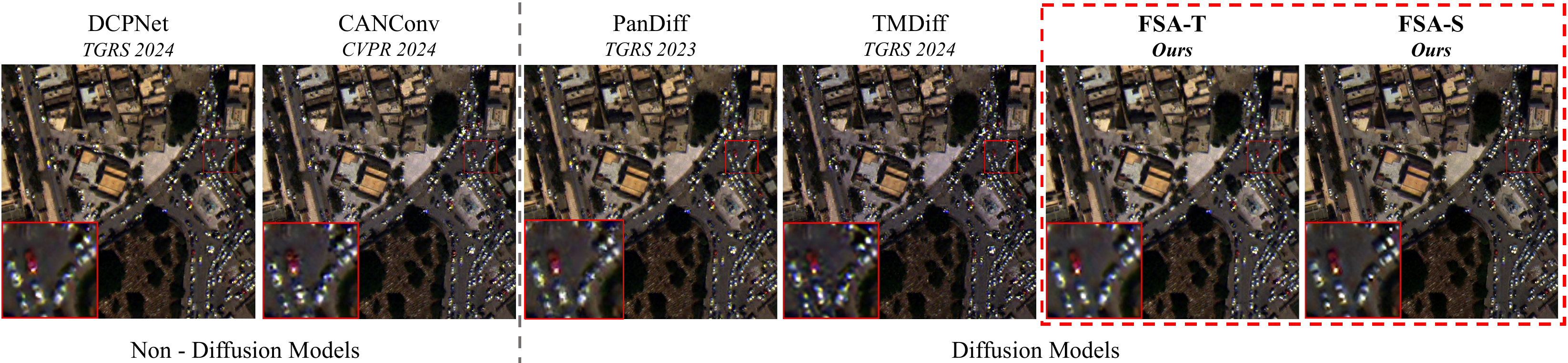}}
    \caption[FR_WV3]{PAN-sharpening results for the WV3 dataset under full resolution (FR) scenarios. The first row depicts the output HRMS images. Zoom in for better visualization.}
    \label{Supple/FR_WV3}
\end{figure*}

\begin{figure*}[tbp]
    \centering{\includegraphics[width=0.99\textwidth]{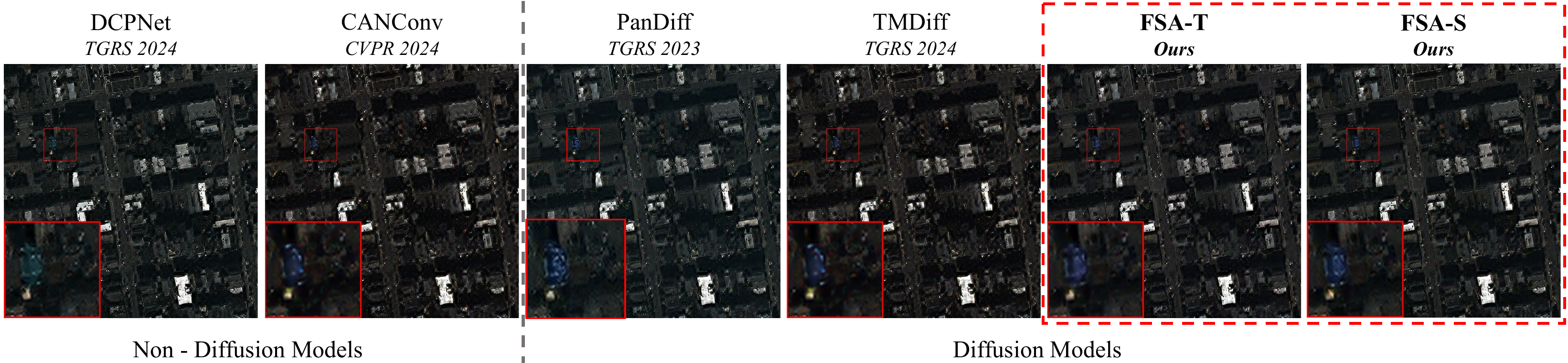}}
    \caption[FR_QB]{PAN-sharpening results for the QB dataset under full resolution (FR) scenarios. The first row depicts the output HRMS images. Zoom in for better visualization.}
    \label{Supple/FR_QB}
\end{figure*}

\begin{figure*}[tbp]
    \centering{\includegraphics[width=0.99\textwidth]{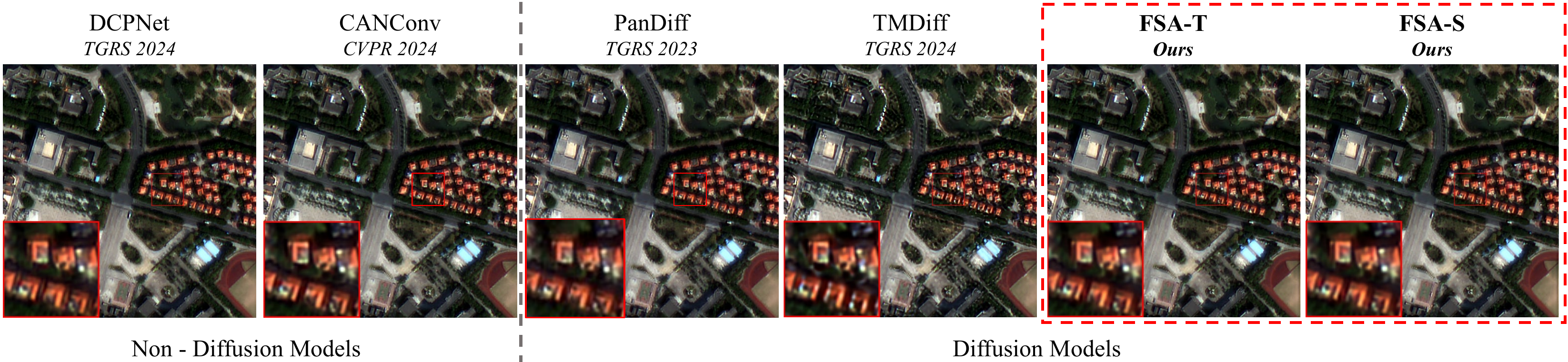}}
    \caption[FR_GF2]{PAN-sharpening results for the GF2 dataset under full resolution (FR) scenarios. The first row depicts the output HRMS images. Zoom in for better visualization.}
    \label{Supple/FR_GF2}
\end{figure*}

Table~\ref{table:WV3,QB,GF2 total result} presents the results of our proposed models, FSA-T and FSA-S, compared with all baseline models on the WV3, QB, and GF2 datasets. The evaluation includes Reduced-Resolution (RR), Full-Resolution (FR), and standard deviation. 
Fig.~\ref{Supple/RR_WV3} to \ref{Supple/FR_GF2} illustrate the qualitative results for the WV3, QB, and GF2 datasets in both Reduced-Resolution and Full-Resolution settings. For the RR results, we visualize the RGB outputs, along with the difference between the output HRMS $\hat{\textbf{I}}^{\text{HR}}_{\text{MS}}$ and the ground truth $\textbf{I}^\text{HR}_\text{MS}$ using error maps and their corresponding mean absolute error (MAE) values. For the FR results, we showcase visual comparisons with the latest state-of-the-art methods. From the visual comparisons, we observe that our U-Know-DiffPAN framework significantly enhances restoration qualities, particularly in the regions with high-frequency contents, high uncertainty, and complex textures, such as edges and small objects. In RR scenarios, these regions exhibit a closer resemblance to their ground truths, compared to the previous methods. Even in FR scenarios where ground truth is unavailable, our U-Know-DiffPAN framework produces more detailed and robust results, demonstrating superior structural and spectral fidelity in high-uncertainty regions. This highlights the capability of our U-Know-DiffPAN to outperform state-of-the-art models in both qualitative and quantitative aspects.

\clearpage
{
    \small
    \bibliographystyle{ieeenat_fullname}
    \bibliography{main}
}
\end{document}